\pdfoutput=1
\documentclass[11pt]{article}

\usepackage{ACL2023}
\usepackage{amssymb}
\usepackage{times}
\usepackage{latexsym}
\usepackage{graphicx}
\usepackage{multirow}
\usepackage[T1]{fontenc}
\usepackage[utf8]{inputenc}
 \usepackage{longtable}
\usepackage{microtype}

\usepackage{inconsolata}

\title{\textit{FinTextQA}: A Dataset for Long-form Financial Question Answering}

\author{Jian Chen\textsuperscript{1 2}\quad{\bf Peilin Zhou\textsuperscript{2}} \quad{\bf Yining Hua\textsuperscript{3}}  \ \\  \ {\bf Yingxin Loh\textsuperscript{1}} \ {\bf Kehui Chen\textsuperscript{1}} {\bf Ziyuan Li\textsuperscript{1}} \bf{Bing Zhu}\textsuperscript{1*} \ {\bf Junwei Liang}\textsuperscript{2}\thanks{*Corresponding author}\\
        \textsuperscript{1}HSBC Lab \quad \textsuperscript{2}Hong Kong University of Science and Technology (Guangzhou)\\
        \textsuperscript{3}Harvard University\\
        \{alex.j.chen, bing1.zhu\}@hsbc.com, \\
        jchen524@connect.hkust-gz.edu.cn, junweiliang@hkust-gz.edu.cn}

\begin{document}
\maketitle
\begin{abstract}
Accurate evaluation of financial question-answering (QA) systems necessitates a comprehensive dataset encompassing diverse question types and contexts. However, current financial QA datasets lack scope diversity and question complexity. This work introduces \textit{FinTextQA}, a novel dataset for long-form question answering (LFQA) in finance.  \textit{FinTextQA} comprises 1,262 high-quality, source-attributed QA pairs extracted and selected from finance textbooks and government agency websites.Moreover, we developed a Retrieval-Augmented Generation (RAG)-based LFQA system, comprising an embedder, retriever, reranker, and generator. A multi-faceted evaluation approach, including human ranking, automatic metrics, and GPT-4 scoring, was employed to benchmark the performance of different LFQA system configurations under heightened noisy conditions. The results indicate that: (1) Among all compared generators, Baichuan2-7B competes closely with GPT-3.5-turbo in accuracy score; (2) The most effective system configuration on our dataset involved setting the embedder, retriever, reranker, and generator as Ada2, Automated Merged Retrieval, Bge-Reranker-Base, and Baichuan2-7B, respectively; (3) models are less susceptible to noise after the length of contexts reaching a specific threshold.
\end{abstract}

\section{Introduction}
The growing demand for financial data analysis and management has led to the expansion of artificial intelligence (AI)-driven question-answering (QA) systems. These systems not only enhance customer service but also assist in risk management and personalized stock recommendations \cite{wu2023ai}. The intricate nature of financial data, with its domain-specific terminologies, concepts, and the inherent uncertainty of the market and decision-making processes, demands a deep understanding of the financial domain to generate accurate and informative responses \citep{confalonieri2021historical}. In this context, long-form question answering (LFQA) scenarios become particularly relevant as they require models to demonstrate a broad spectrum of sophisticated skills, including information retrieval, summarization, data analysis, comprehension, and reasoning \citep{fan2019eli5}.

\begin{figure}[t]
    \centering
    \includegraphics[width=0.8\linewidth]{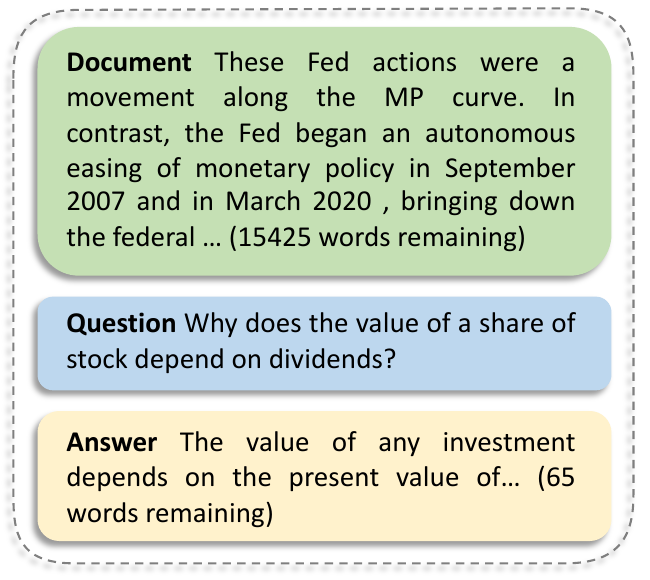}
    \caption{An LFQA sample in \textit{FinTextQA}. Models are expected to generate paragraph-length answers when given questions and documents.}
    \label{fig:lfqa_sample}
\end{figure}

\begin{figure*}[t]
    \centering
    \includegraphics[width=1\linewidth]{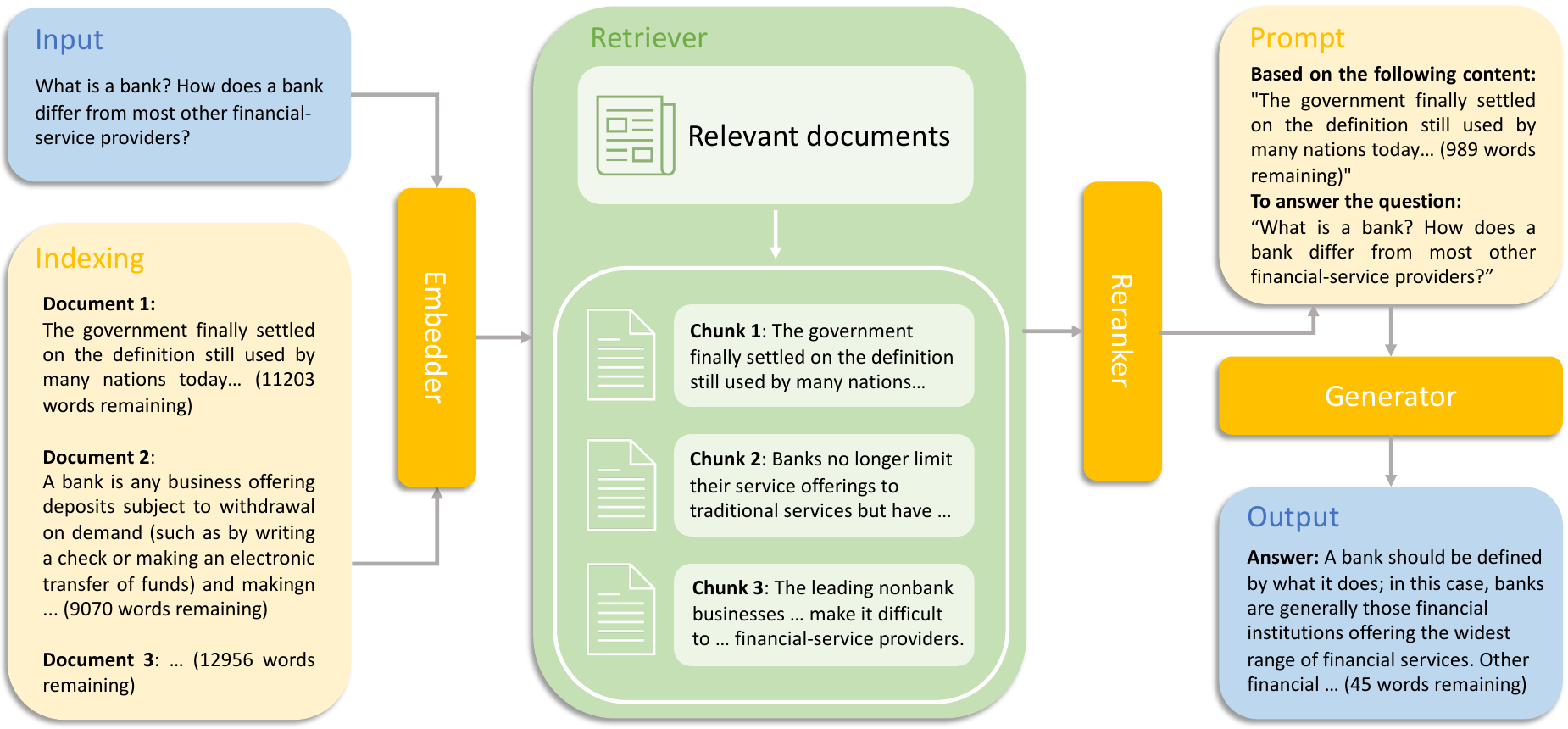}
    \caption{The workflow of our proposed RAG-based LFQA system. Embedder aims to encode documents and user's question into semantic vectors. Retriever retrieves relevant document chunks based on the encoded question. Reranker removes less-similar chunks. With a prompt which combines question and chunks, the generator finally output desired answer.}
    \label{fig:rag_process}
\end{figure*}

In the general domain, there are several LFQA datasets available, including ELI5 \citep{fan2019eli5}, WikiHowQA \citep{bolotova2023wikihowqa} and WebCPM \citep{qin2023webcpm}. However, it is important to note that there is currently no LFQA dataset specifically tailored for the finance domain. Existing financial QA benchmarks often fall short in addressing question complexity and variety by primarily on sentiment analysis and numerical calculation, as comprehensive paragraph-length responses and relevant document retrievals are often required to answer intricate, open-domain questions \citep{han2023prewome}. To address these challenges, we introduce a new dataset, \textit{FinTextQA}, which comprises LFQAs from finance-related textbooks and government agency websites to assess QA models on general finance and regulation or policy-related questions. \textit{FinTextQA} consists of 1,262 high-quality, source-attributed question-answer pairs and associated document contexts. It contains six question types with an average text length of 19.7k words, curated from five rounds of human screening. This dataset is pioneering work in integrating financial regulations and policies into LFQA, challenging models with more demanding content.

In addition to introducing the dataset, we conduct comprehensive benchmarking of state-of-the-art (\textit{sota}) models on \textit{FinTextQA} to provide baselines for future research. Current LFQA systems frequently solely rely on fine-tuning pre-trained language models such as GPT-3.5-turbo, LLaMA2 \cite{touvron2023llama}, Baichuan2 \cite{yang2023baichuan}, etc., which often fail to provide detailed explanations or effectively handling complicated finance questions \cite{yuan2023evaluating}. In response, we opt for the Retrieval-augmented generation (RAG) framework, as illustrated in Figure \ref{fig:rag_process}. By processing documents in multiple steps,  RAG systems can pre-process and provide the most relevant information to LLMs, enhancing their performance and explanation capabilities \cite{guu2020retrieval}. 

We believe this work, by introducing the first LFQA financial dataset and conducting comprehensive benchmark experiments on the dataset, marks a milestone in advancing the comprehension of financial concepts and enhancing assistance in this field: \textit{FinTextQA} offers a rich and rigorous framework for building and assessing the capabilities of general finance LFQA systems. Our experimental analysis not only highlights the efficacy of various model configurations but also underscores the critical need for enhancing current methodologies to improve both the precision and explicability of financial question-answering systems.

\section{Related Work}
\subsection{Long-Form Question Answering (LFQA)}
The goal of LFQA is to generate comprehensive, paragraph-length responses by retrieving and assimilating relevant information from various sources \citep{fan2019eli5}. This poses a significant test for current Natural Language Processing (NLP) and Artificial Intelligence (AI) models, given their limited understanding and learning capacities \citep{thompson2020computational}. 

Several LFQA datasets are available in general domain, including ELI5 \citep{fan2019eli5}, WikiHowQA \citep{bolotova2023wikihowqa}, and WebCPM \citep{qin2023webcpm}. In the financial domain, some QA datasets have been developed. However, none of them addresses LFQA. While these datasets like FinQA \citep{chen2021finqa} and TAT-QA \citep{zhu2021tat} address specific scopes such as numerical reasoning, they do not touch upon general LFQA tasks. In addition, FIQA \citep{maia201818} only provides short-context documents, which may not adequately represent real-life scenarios and have limited industry applicability. Although FinanceBench \citep{islam2023financebench} does cover a wider scope, it only offers 150 open-source question-answer pairs, while the question complexity and answer do not satisfy real-life LFQA scenarios. 

\subsection{Retrieval-Augmented Generation (RAG)}
% One key challenge is improving sample efficiency for training large models. Without more efficient methods, tackling complex LFQA tasks may become economically unviable. Another issue is shortcut learning \citep{geirhos2020shortcut}, where AI models rely on spurious statistical cues for test performance but often struggle to generalize to new data, hindering overall performance and applicability.

% In response, a branch of research has adopted a three-fold approach, fusing retrieval, text generation, and open-domain QA methods - RAG \citep{krishna2021hurdles, su2022read}. By assimilating short-answer techniques like ORQA \citep{tchakaloff2013orqa}, REALM \citep{guu2020retrieval}, and DPR \citep{karpukhin2020dense} into pre-trained language models, they are shown to effectively address complex tasks \cite{fan2019eli5}. This interdisciplinary methodology works effectively for diverse question types, from factoid to non-factoid, including multi-hop reasoning across different paragraphs \citep{bui2020state}.

RAG frameworks represent a significant advancement in LFQA, incorporating external knowledge sources and In-Context Learning (ICL) for efficient information retrieval and application. By combining diverse documents into comprehensive prompts, RAG enables language models to generate contextually informed responses without task-specific retraining \citep{lewis2020retrieval}. 

The evolution of RAG involves three principal stages: Naive RAG, Advanced RAG, and Modular RAG. Naive RAG offers improvements over traditional language models by providing cost-efficient indexing, retrieval, and generation, albeit with certain constraints. Advanced RAG addresses these limitations by integrating refined indexing and retrieval techniques, optimizing data handling, and introducing strategic Retrieval and post-retrieval processes. Its capabilities include fine-tuning domain-specific embedders, employing dynamic ones for improved context comprehension, and applying reranker and prompt compression during post-retrieval processes \citep{ilin2023advanced}. Modular RAG represents a further advancement from traditional NLP frameworks by introducing specialized modules for similarity-based retrieval, fine-tuning, and problem-solving. It also incorporates innovative modules like Search and Memory \citep{cheng2023lift}, Fusion \citep{rackauckas2024rag}, and Routing to customize RAG for specific applications and improve search and retrieval operation \citep{gao2023retrieval}. The ongoing evolution of RAG demonstrates its potential to revolutionize information retrieval and adaptability in language model systems within the fast-evolving field of computational linguistics. In this study, we choose to assess the effectiveness of Modular RAG with the rewrite, retrieve, re-rank, and read modules following previous work \cite{gao2023retrieval}.

\section{\textit{The FinTextQA Dataset}}
\subsection{Data Sources}
The data in \textit{FinTextQA} are sourced from well-established financial literature and government agencies, such as expert-authored question-answer pairs from recognized finance textbooks: \textit{Bank Management and Financial Services} (BMFS), \textit{Fundamentals of Corporate Finance} (FCF), and \textit{The Economics of Money, Banking, and Financial Markets} (EMBFM). Additionally, crucial information regarding financial regulations and policies is incorporated from esteemed websites such as the Hong Kong Monetary Authority (HKMA) \footnote{\url{https://www.hkma.gov.hk/eng}}, European Union (EU)\footnote{\url{https://european-union.europa.eu/index_en}}, and the Federal Reserve (FR)\footnote{\url{https://www.federalreserve.gov}}.  Question types encompass various domains, spanning concept explanation and numerical calculation to comparative analysis and open-ended opinion-based queries.

% With our expert-conducted data selection and annotation, FinText ensures quality, credibility, and comprehensiveness.

% Given the well-established quality of finance textbooks and their ongoing assessment within the academic sphere, we directed our attention towards assessing the policy and regulation dataset. This dataset is unique due to its inherent complexity; the questions and answers may not always have a direct correspondence, often necessitating a reasoning process to infer the appropriate responses. To address this challenge and guarantee the quality of the dataset, we designed a stringent, two-step verification process:

\subsection{Selection of Policy and Regulation Data}
In textbooks, questions are typically straightforward, and evidence (i.e., citations) can be easily found within each chapter. However, in policies and regulations, some of the questions draw from multiple sources and may not directly align with the documents in our dataset. This poses a challenge for the model to provide accurate answers. Additionally, policy and regulation data often require deeper analytical thinking and interpretation, demanding a robust reasoning ability from the QA system. Given the complexity and importance of financial regulations and policies, we have implemented a thorough two-step verification process to ensure the relevance and accuracy of the QA pairs and the associated regulation and policy documents:

\begin{enumerate}
    \item \textbf{Evidence identification}: Initially, annotators are tasked with locating relevant evidence (aka citations and references) for each question-answer pair within the dataset. Any questions that cannot be feasibly linked to a valid citation or reference were promptly excluded from consideration; 
    \item \textbf{Relevance evaluation}: Another distinct group of annotators evaluates the coherence and connectedness between the question, context, and answer for each entry. Using a grading scale from 1 to 5, they ensure high standards of relevancy. Only entries with a score exceeding 2 across all three variables are included in the final dataset.
\end{enumerate}

% In total, we have 1322 question-answer pairs with related docment context in the dataset, and 1262 of which were retained after pre-processing. This rigorous process produced a dataset displaying strong relevance among answer-context (3.91), question-answer (4.88), and question-context (4.54), indicative of high quality and dependability. The refined dataset allows comprehensive evaluation of AI models in terms of their performance in finance-related tasks using FinTextQA.

Initially, we collected 300 regulation and policy question-answer pairs with related document contexts. After careful data quality control, 240 pairs were retained. The data selection process resulted in a dataset demonstrating strong relevance among answer-context (3.91), question-answer (4.88), and question-context (4.54), indicating its high quality and dependability. Further details of human evaluation can be found in Appendix \ref{sec:appendix}.

\begin{table}[t]
\centering

\setlength{\tabcolsep}{5mm}{%
\begin{tabular}{lcc}

\hline
\textbf{Source}   &\textbf{\begin{tabular}[c]{@{}c@{}}$\#$ of \\Document\end{tabular}}&\textbf{\begin{tabular}[c]{@{}c@{}}$\#$ of \\Question\end{tabular}} \\\hline
EU        &1                & 12 \\
FR       &8                & 190 \\
HKMA                  &6                & 38 \\
{\begin{tabular}[l]{@{}l@{}}BMFS\end{tabular}}     &19        & 319 \\
{\begin{tabular}[l]{@{}l@{}}EMBFM\end{tabular}}    &26        & 472 \\
FCF                   &20               & 231 \\\hline
\end{tabular}}
\caption{Disribution of numbers of documents and questions from different sources.}
\label{tab:doc_query_source}
\end{table}

\subsection{Dataset Statistics}
\textit{FinTextQA} contains 1,262 QA pairs, with 1,022 pairs from finance textbooks, accounting for 80.98\% of the dataset, and 240 pairs from policies and regulations, accounting for 19.02\% of the dataset. We randomly split the dataset into training, validation, and test sets following a 7:1:2 ratio for model fine-tuning and evaluation. Table \ref{tab:doc_query_source} presents data distribution across different sources.

Table \ref{tab:doc_query_source} illustrates the distribution of these questions, representing various aspects of financial regulations and policies. The European Commission subset comprises 12 questions focused on transaction regulation and its interpretations. 
% An example question is: "Under which circumstances may a financial contribution provided by a private entity be attributed to a third country under Article 3(3)(c) of Regulation (EU) 2022/2560?". 
The Federal Reserve subset, containing over 190 questions, addresses topics such as banking regulations, monetary policy strategies, and international banking operations.
% For instance: "In what ways may a U.S. banking organization conduct operations abroad under Regulation K?". 
The Hong Kong Monetary Authority subset contains 38 questions covering anti-money laundering, counter-terrorist financing ordinance, and credit card business regulations, etc.
% , with questions like "Can a bank close my account?".

% Furthermore, the BMFS, EMBFM, and FCF datasets collectively contain 1,022 questions pertaining to various aspects of banking and finance. These questions encompass both concept-based and open-ended queries, such as "Under U.S. law, what must a corporation do to qualify and be regulated as a commercial bank?".

\begin{figure}[t]
    \centering
    \includegraphics[width=1\linewidth]{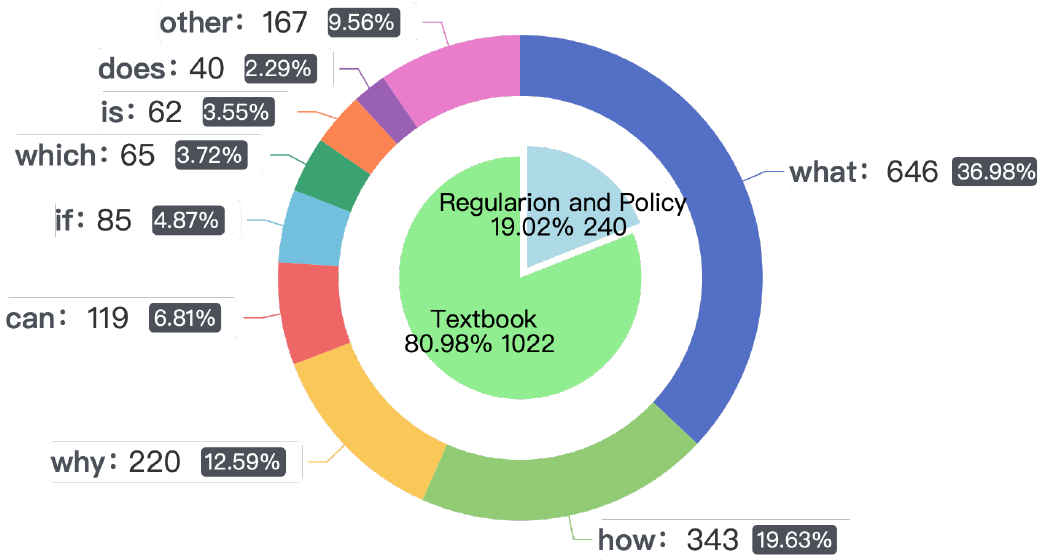}
    \caption{Distribution of data sources and interrogative words in  \textit{FinTextQA}.}
    \label{fig:question_type_distribution}
\end{figure}

% The questions in this dataset are primarily compound, with each main question encompassing 2-3 sub-questions. These sub-questions display various types, contributing to a diverse range of interrogative words, as shown in Figure \ref{fig:fig2}. We observe that 36.98$\%$ of questions began with "what", followed by 19.63$\%$ with "how", 12.59$\%$ with "why", and 6.81$\%$ with "can". This diverse question composition allows for a more comprehensive evaluation of the reading and understanding capabilities of large language models.
\textit{FinTextQA} consists mainly of compound questions, where each primary question includes 2-3 related sub-questions. 
This hierarchical format introduces more complexity for question understanding and reasoning.
These sub-questions come in different forms, leading to a variety of interrogative words as illustrated in Figure \ref{fig:question_type_distribution}. Our analysis shows that 36.98\% of the questions start with "what", making it the most common starting word, followed by "how" at 19.63\%, "why" at 12.59\%, and "can" at 6.81\%. The diversity in the types of interrogative words enriches the dataset, providing a more thorough test of large language models' ability to read and understand text.

\subsection{Comparison to Existing Datasets}
Table \ref{tab:dataset_comparison} shows a comparison of LFQA datasets, not limited to finance. \textit{FinTextQA} stands out with an average question length of 28.5 words, answers of 75 words, and notably extended document contexts, averaging 19,779.5 words. These extensive contexts, segmented into chapters or sessions, are designed to enhance retrieval tasks. Furthermore, \textit{FinTextQA} covers a broad scope, including multi-turn, numerical, finance domain, and open-ended questions. It contains the most complex questions and longest answers alongside the widest scope, as compared with other finance QA datasets. Further details of question types can be found in Appendix \ref{sec:appendixqt}.

\begin{table*}[t]
\centering
\tiny
\renewcommand{\arraystretch}{1.2}
\resizebox{1.0\textwidth}{!}{%
\begin{tabular}{c|ccc|cccccc}

\hline
\multirow{2}{*}{\textbf{Dataset}} & \multicolumn{3}{c|}{\textbf{Average $\#$ of Words}}  & \multicolumn{6}{c}{\textbf{Scope}}      \\
& \textbf{Question} & \textbf{Document} & \textbf{Answer}& \textbf{Multi-turn} & \textbf{Comparative} & \textbf{Numerical}& \textbf{Domain} & \textbf{Open-minded} & \textbf{Cause and Effect}\\ \hline
FIQA \cite{maia201818}& 12.8& 136.4 & - & && &\checkmark &\checkmark&\\
TAT-QA \cite{zhu2021tat}& 12.4& 42.6& 4.3& & &\checkmark &\checkmark &&\\
FinQA \cite{chen2021finqa}& 16.6& 628.1& 1.1&\checkmark & \checkmark&\checkmark & \checkmark& &\\
FinanceBench \cite{islam2023financebench}& 27.0& \textbf{65,615.6}& 12.66 & & &\checkmark &\checkmark &\checkmark       &  \\ \hline
\textbf{ FinTextQA (ours)}& \textbf{28.5}& 19,779.5& \textbf{75}&\checkmark &\checkmark &\checkmark &\checkmark &\checkmark &\checkmark           \\ \hline
\end{tabular}}
\caption{Comparison of various financial QA datasets. \textit{FinTextQA} offers substantially longer questions and answers. Meanwhile, has a wider scope compared with other finance QA datasets.}
\label{tab:dataset_comparison}
\end{table*}

\begin{table*}[t]
\tiny
\centering
\renewcommand{\arraystretch}{1.1} 
\resizebox{1.0\textwidth}{!}{%
\begin{tabular}{llllccccc}
\hline

\textbf{Generator} & \textbf{Retriever} & \textbf{Embedder} & \textbf{Reranker} & \textbf{\begin{tabular}[c]{@{}c@{}}Answer \\ Accuracy\end{tabular}} & \textbf{ROUGE-1} & \textbf{ROUGE-2} & \textbf{ROUGE-L} & \textbf{BLEU} \\  \hline
\multirow{3}{*}{GPT-3.5-turbo} & AMR & Ada2 & LLMRerank & 4.411 & \textbf{0.346} & \textbf{0.134} & \textbf{0.224} & \textbf{0.062} \\
& AMR & Ember-v1 & LLMRerank & 4.365 & 0.341 & 0.130 & 0.221 & 0.060 \\
& AMR & Ember-v1 & Bge-Reranker-Base & 4.439 & 0.339 & 0.131 & 0.221 & 0.062 \\  \hline

\multirow{3}{*}{Baichuan2-7B} & AMR & Ada2 & LLMRerank & 4.578 & 0.340 & 0.124 & 0.219 & 0.057 \\
& AMR & Ada2 & Bge-Reranker-Base & \textbf{4.612} & 0.338 & 0.123 & 0.217 & 0.054 \\
& AMR & Ember-v1 & Bge-Reranker-Base & 4.513 & 0.333 & 0.120 & 0.215 & 0.053 \\  \hline

\multirow{3}{*}{Solar-10.7B} & AMR & Ember-v1 & Bge-Reranker-Base & 4.348 & 0.329 & 0.119 & 0.205 & 0.052 \\
& AMR & Bge-Small-en-v1.5 & Bge-Reranker-Base & 4.310 & 0.329 & 0.118 & 0.205 & 0.051 \\
& AMR & Ada2 & Bge-Reranker-Base & 4.378 & 0.327 & 0.119 & 0.204 & 0.051 \\      \hline

\multirow{3}{*}{Qwen-7B} & AMR & Bge-Small-en-v1.5 & LLMRerank & 4.414 & 0.341 & 0.125 & 0.217 & 0.059 \\
& AMR & Ada2 & Bge-Reranker-Base & 4.405 & 0.337 & 0.120 & 0.216 & 0.056 \\
& AMR & Ember-v1 & LLMRerank & 4.432 & 0.339 & 0.121 & 0.215 & 0.056 \\          \hline

\multirow{3}{*}{LLaMA2-7B} & SWR & Ada2 & All-Mpnet-Base-v2 & 4.184 & 0.233 & 0.078 & 0.152 & 0.030 \\
& AMR & Bge-Small-en-v1.5 & Bge-Reranker-Base & 4.268 & 0.239 & 0.078 & 0.151 & 0.031 \\
& AMR & Bge-Small-en-v1.5 & LLMRerank & 4.287 & 0.233 & 0.076 & 0.149 & 0.031 \\ \hline

\multirow{3}{*}{Gemini-Pro} & AMR & Ember-v1 & Bge-Reranker-Base & 3.970 & 0.304 & 0.118 & 0.211 & 0.048 \\
& AMR & Ember-v1 & LLMRerank & 3.990 & 0.306 & 0.119 & 0.211 & 0.052 \\
& AMR & Bge-Small-en-v1.5 & LLMRerank & 3.989 & 0.303 & 0.119 & 0.210 & 0.051 \\ \hline
\end{tabular}}
\caption{Systematic performance comparison of RAG-based LFQA system with different configurations.}
\label{tab:result_rag_system}
\end{table*}

\begin{table}[t]
\small
\centering
\renewcommand{\arraystretch}{1.1} 
\resizebox{1.0\columnwidth}{!}{%

\begin{tabular}{llc}
\hline
\textbf{Module}            & \multicolumn{1}{l}{\textbf{Model}}                      &  \textbf{\begin{tabular}[c]{@{}c@{}}Average \\Score\end{tabular}}              \\ 
\hline
\multirow{4}{*}{Embedder}    & Ada2                                                    & \textbf{4.586} \\
                              & Ember-v1                                                & 4.486 \\
                              & Bge-Small-en-v1.5                                       & 4.455 \\
                              & Gte-Large                                               & 4.261 \\ \hline
\multirow{3}{*}{Retriever}    & AMR                                  & \textbf{4.492} \\
                              & SWR                               & 4.466 \\
                              & Vector Retrieval                                        & 4.358 \\ \hline
\multirow{3}{*}{Reranker}     & Bge-Reranker-Base                                       & \textbf{4.489} \\
                              & LLMRerank                                 & 4.469 \\
                              & All-Mpnet-Base-v2                                       & 4.383 \\ \hline
\multirow{3}{*}{\begin{tabular}[l]{@{}l@{}}Best-perfoming\\Configurations\end{tabular}}  & AMR+Ada2+LLMRerank      & \textbf{4.622} \\
                              & SWR+Ada2+Bge-Reranker-Base      & 4.620 \\
                              & SWR+Ada2+All-Mpnet-Base-v2      & 4.620
                              \\ \hline
\end{tabular}}
\caption{Performance comparison of embedders, retrievers, rerankers, and best-performing configuration. GPT-4 scores are generated regarding question-evidence relevance.}
\label{tab:result_evidence_framework}
\end{table}

\begin{table*}[t]
    \centering
    \small
    \renewcommand{\arraystretch}{1.1} 
    \setlength{\tabcolsep}{1.7mm}{\begin{tabular}{lcccccccccc}
    \hline
    \multirow{2}{*}{\textbf{Model}}
    &  \multicolumn{2}{c}{\textbf{\begin{tabular}[c]{@{}c@{}} $\#$ of Unanswered \\Questions\end{tabular}} }
         &  \multicolumn{2}{c}{\textbf{\begin{tabular}[c]{@{}c@{}}Accuracy(\%)\end{tabular}}}
         &  \multicolumn{2}{c}{\textbf{\begin{tabular}[c]{@{}c@{}}Answer\&Evidence\\Relevance\end{tabular}}}
         &  \multicolumn{2}{c}{\textbf{ROUGE-L}}
         &  \multicolumn{2}{c}{\textbf{BLEU}}\\ \cline{2-11}
         
         &  Base&  Fine-tuned&  Base&  Fine-tuned&  Base&  Fine-tuned&  Base&  Fine-tuned&  Base&Fine-tuned\\ 
\hline
         GPT-3.5-turbo                         
         &  21&  \textbf{0}&  4.30&  4.08&  4.34&  4.39&  \textbf{0.21}&  0.19&  \textbf{0.05}&0.03
\\ 
         Baichuan2-7B                          
         &  \textbf{0}&  \textbf{0}&  \textbf{4.50}&  \textbf{4.51}&  \textbf{4.73}&  \textbf{4.73}&  0.20&  \textbf{0.20}&  \textbf{0.05}&\textbf{0.04}
\\ 
         Qwen-7B                          
         &  13&  10&  4.43&  4.43&  4.59&  4.35&  0.19&  0.19&  0.04&\textbf{0.04}
\\ 
         Solar-10.7B                        
         &  13&  12&  4.38&  4.38&  4.50&  4.50&  0.19&  0.18&  0.04&\textbf{0.04}
\\ 
         LLaMA2-7B                       
         &  \textbf{0}&  \textbf{0}&  4.14&  4.27&  4.22&  4.28&  0.10&  0.10&  0.02&0.02
\\ 
         Gemini-Pro                         
         &  61&-  &  2.46&-  &  1.85&-  &  0.15&-  &  0.02&-\\ \hline
    \end{tabular}}
    \caption{Performance comparison of generators. \# of Unanswered Questions is the number of "can not provide answer based on the content" generated by different generators. We use the same embedder, retriever, and reranker in this experiment. An example of unanswered questions is shown in Appendix Table \ref{tab:uaq}.}
    \label{tab:generator_result}
\end{table*}
% \subsection{Dataset Characteristics}
% \noindent\textbf{Reliability and Authenticity\quad} 's reliability and authenticity stem from its foundation in established financial literature and authoritative government sources. By leveraging expert-authored question-answer pairs from recognized finance textbooks, the dataset ensures that the content is both accurate and relevant. Additionally, the incorporation of key financial regulations and policies from renowned government agencies guarantees domain-specific credibility. The two-step verification process (citation and reference identification, and relevance evaluation) bolsters reliability. The well-established reputation of finance textbooks, combined with the ongoing assessment within the academic sphere, adds to FinText's robustness.
 
% \noindent\textbf{Standardization and Comprehensiveness\quad} The dataset maintains a consistent system by segmenting the information into chapters or sessions. It comprises a comprehensive array of question-answer pairs from finance textbooks and regulatory and policy data. The variety of question types and coverage of diverse topics within the financial domain contribute to its comprehensiveness. The dataset allows for a thorough evaluation of AI models' performance in finance-related tasks.

\section{Benchmarks on \textit{FinTextQA}}
\subsection{RAG-based LFQA system}
We employ the modular RAG as discussed in \cite{gao2023retrieval} and follow the guidelines outlined in LlamaIndex\footnote{\url{https://www.llamaindex.ai}} to construct the RAG-based LFQA system. As shown in Figure \ref{fig:rag_process}, this LFQA system consists of four modules: embedder, retriever, reranker, and generator. The first three modules together serve to find relevant information (aka \textit{evidence} or \textit{citations}) from contexts. The last module synthesizes responses using the retrieved information. Each module can be implemented by different models, and the combinations of models for all modules constitutes the system's configurations. The selection of the models for each module is summarzied as follows:

\noindent\textbf{The Embedder Module\quad}
The role of the embedder module is to convert human language into a vector representation that can be understood and processed by computers. In our experiments, we adopt four popular embedding models that have achieved high rankings on the Hugging Face leaderboard, including (1) BAAI's Bge-small-en-v1.5 \citep{Xiao_Liu_2023a}, (2) NLPer's Gte-large \cite{li2023towards}, (3) LLMRails' Ember-v1\footnote{\url{https://huggingface.co/llmrails/ember-v1}}, and (4) OpenAI's Ada2\footnote{\url{https://platform.openai.com/docs/guides/embeddings}}.

\noindent\textbf{The Retriever Module\quad}
The retriever module forms the backbone of our experiment by searching and retrieving relevant context related to a given question. We explore three retriever methods, including Auto Merging Retriever (AMR) \citep{Liu_2023}, (2) Sentence Window Retriever (SWR) \citep{llamaindex_2023a}, and a simple vector-based retriever approach. AMR organizes documents into a hierarchical tree system with parent nodes' contents distributed among child nodes. This enables users to determine the relevance of the parent node based on its child nodes' relevance to the query. SWR fetches context from a custom knowledge base by considering a broader context and retrieving sentences around the most relevant sentence. This leads to the generation of higher-quality context. Finally, the vector-based retriever approach simply searches for related context through a vector index.

\noindent\textbf{The Reranker Module\quad}
The primary objective of rerankers is to refine the retrieved information by repositioning the most pertinent content towards the prompt edges. To accomplish this, we examine the influence of three rerankers on the overall system performance: (1) LLMRerank \citep{Fajardo_2023}, (2) Bge-Ranker-Base\footnote{\url{https://huggingface.co/BAAI/bge-reranker-base}}, and (3) All-Mpnet-Base-v2\citep{song2020mpnet}.

\noindent\textbf{The Generator Module\quad}
The generator module first consolidates the query and relevant document context prepared by the former modules into a well-structured and coherent prompt. These prompts are then fed to a LLM to generate final responses. To evaluate the performance of various LLMs, we include six \textit{sota} models, including (1) Qwen-7B \citep{bai2023qwen}, (2) Baichuan2-7B \citep{yang2023baichuan}, (3)LLaMA2-7B \citep{touvron2023llama}, (4) GPT-3.5-turbo, (5) Solar-10.7B \citep{kim2023solar}, and (6) Gemini-Pro \citep{team2023gemini}. 

\subsection{Experimental Settings}
To ensure a thorough understanding of each model within every module in a controlled manner, we systematically tested all configurations of models in each module in the RAG-based LFQA system to determine the optimal one. All configurations are evaluated on two sets of experiments - one where the generators were fine-tuned using the training set of \textit{FinTextQA}, and another without such fine-tuning. Note that Gemini-Pro remains a private model and is thus excluded from the fine-tuning process. 

To understand the robustness of the best systems, we select the three highest-ranking configurations based on their performance with generators in their base form. This criterion ensures a fair comparison with Gemini-Pro. We then evaluate the performance of these systems under conditions of increased noise by incrementally adding numbers of documents from one to three.

Hyperparameter settings involved in the experiments are set as follows:

\noindent\textbf{Retrievers\quad} For AMR, we define three levels of chunk sizes: 2048 for the first level, 512 for the second, and 128 for the third. For the SWR method, we set the window size to 3. For all retrievers, the similarity top \textit{k} value was set to 6.

\noindent\textbf{Rerankers\quad} We set the LLMRerank batch size to 5, and the top \textit{n} values of LLMRerank, Bge-Reranker-Base, and All-Mpnet-Base-v2 to 4.

\noindent\textbf{Generator\quad} We use AzureOpenai's API \footnote{\url{https://azure.microsoft.com/en-us/products/ai-services/openai-service}} to access GPT-3.5-turbo and GPT-4-0314 for GPT series models. Google VertexAI API\footnote{\url{https://cloud.google.com/vertex-ai/docs/reference/rest}} is used to access Gemini-Pro. The LLaMA2, Baichuan2, and Qwen models are all used in their 7B versions, while the Solar model is accessed in the 10.7B version. Fine-tuning of open-source models is carried out on the training set of \textit{FinTextQA}. For GPT-3.5-turbo, we adopt the fine-tuning methods in Azure AI Studio\footnote{\url{https://oai.azure.com}}, setting the batch size to 2, learning rate multiplier to 1, and epochs to 5. 

GPTQConfig \cite{frantar2022gptq} is used to load the Qwen-7B model in 4-bit, the GenerationConfig \citep{Huggingface_2022} for Baichuan2-7B in 4-bit, and the BitsAndBytesConfig \citep{Huggingface_2023} for LLaMA2-7B and Solar-10.7B in 4-bit. We employ LoRA for LLaMA2-7B, Baichuan2-7B, Qwen-7B, and Solar-10.7B, with the rank set to 1, alpha set to 32, and dropout at 0.1. Prefix token lengths are set to 2048, learning rate to 1.0e-3, batch size to 2, and maximum input and target length to 2048. All fine-tuning efforts are performed using 12 NVIDIA RTX3090 GPUs for 10 epochs.

\subsection{Evaluation Methods}
% We propose to use a two-fold evaluation approach for the overall performance of retrieval and post-retrieval models by examining the LFQA tasks performance and long-form answer generation outputs. 
\subsubsection{Evaluation of Individual Modules}
\noindent\textbf{Embedders, Retrievers, and Rerankers.} To evaluate the performance of these modules and their combined performance in evidence generation, we use GPT-4 to analyze the relevance between questions and retrieved citations (aka. evidence). In detail, GPT-4 is asked to grade the question-evidence relevance on a five-point Likert scale. The average score, referred to as the `GPT-4 score', is calculated for overall performance evaluation. The prompt used for GPT-4-aided evaluation is shown in Appendix \ref{sec:appendixprompt}.
% 总分（genral)-helpfulness, relevance, accuracy, depth, and creativity., question-evidence 相关性，evidence-答案相关性

\noindent\textbf{Generators.}
To evaluate the performance of generators, we employ automatic metrics comprising matching-based measures such as ROUGE-1, ROUGE-2, and ROUGE-L \citep{lin2004rouge}, as well as the BLEU score \citep{papineni2002bleu}. However, since prior research \citep{zheng2023judging} shows that matching-based metrics may overestimate performance in long sequences, we also use the GPT-4 evaluation method mentioned above to assess evidence-answer relevance. In addition, we report the ratio of unanswered questions in the responses (e.g., cases when models return "can not provide answer based on the content).

\subsubsection{Overall Evaluation of the RAG-based LFQA System} 
ROUGE \citep{lin2004rouge} and BLEU \citep{papineni2002bleu}) are used to automatically measure the overall system performance. The GPT-4 scoring method is used to evaluate the answers from helpfulness, relevance, accuracy, depth, and creativity. Additionally, we invite three annotators to rank top-performing answers from all tested models and compare them with the ground truth answers, capturing human perception and assessing subjective response quality. Further details of human evaluation can be found in Appendix \ref{sec:appendix}. 

\begin{figure*}[t]
    \centering
    \includegraphics[width=1\linewidth]{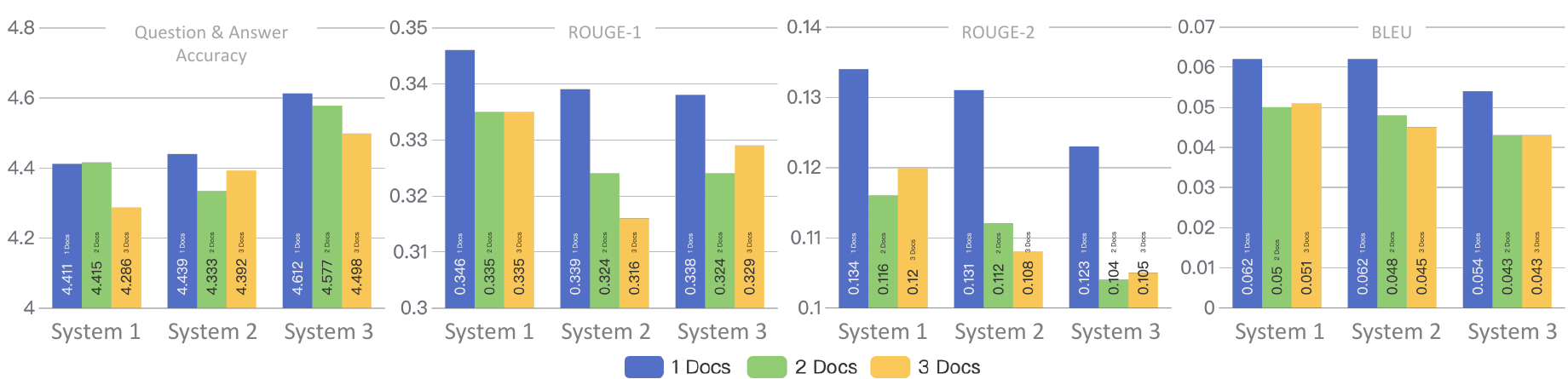}
    \caption{Evaluation of Three Best-performing system configurations with Different Numbers of Input Documents}
    \label{fig:result_noise}
\end{figure*}

\subsection{Results}

\noindent\textbf{Embedders, Retrievers, and Rerankers\quad}
\noindent Table \ref{tab:result_evidence_framework} shows the GPT-4 score of different embedders, retrievers, and rerankers, which constitute the evidence generation pipeline. It also shows at the end the best-performing evidence-generation module combinations. We observe that the highest-performing embedding model is Ada2, achieving a score of 4.586, followed by Ember-v1 (4.486) and Bge-Small-en-v1.5 (4.455) with similar scores. Gte-Large lagged with a noticeable gap with a score of 4.261. 
Among the retrievers we assess, AMR outperforms the rest with an impressive score of 4.492. SWR ranks second at 4.466, while the simple vector-based approach has the lowest performance with a score of 4.358. 

Among the rerankers, Bge-Reranker-Base performs the best, achieving a competitive score of 4.489. LLMRerank ranks second with a score of 4.469, followed by All-Mpnet-Base-v2 with a score of 4.383.

The evidence generation modules together, we observe that the combination of AMR, Ada2, and Bge-Reranker-Base yields the highest score of 4.622, followed by the combination of SWR, Ada2, and Bge-Reranker-Base/All-Mpnet-Base-v2, with a score of 4.620. The marginal differences in performance among these leading combinations indicate that a variety of configurations are capable of yielding satisfactory outcomes for evidence generation.

\begin{table}[t]
\centering
\small
\renewcommand{\arraystretch}{1}
\begin{tabular}{lccccc}

\hline
 \textbf{}   &\textbf{\begin{tabular}[c]{@{}c@{}}Answer\\One\end{tabular}}  &\textbf{\begin{tabular}[c]{@{}c@{}}Answer\\Two\end{tabular}}  &\textbf{\begin{tabular}[c]{@{}c@{}}Answer\\Three\end{tabular}}
 &\textbf{\begin{tabular}[c]{@{}c@{}}Answer\\Four\end{tabular}}\\\hline
\textbf{\begin{tabular}[c]{@{}c@{}}Average\\Ranking\end{tabular}} & 2.19 &2.11	&3.10	&2.60 \\\hline
\end{tabular}
\caption{Comparison of the average rankings of four answers generated by top RAG systems.}
\label{tab:result_human_ranking}
\end{table}

\noindent\textbf{Generators\quad}
Table \ref{tab:generator_result} shows the comparison of different generators, contrasted by their base form and fine-tuned form. Although the fine-tuned models have a decreasing loss (from 2.5 to 0.1), they do not have significant improvement. They demonstrate slightly lower performance in terms of GPT-4 score, ROUGE-L, and BLEU scores. However, fine-tuned models have less unanswered questions, showing better understanding capabilities than their base forms. 

 We also observe that while Gemini-Pro shows high numeric scores, it struggles the most in generating contextually relevant responses. Conversely, Baichuan2-7B demonstrates the best prompt comprehension ability. GPT-3.5-turbo experiences more difficulty with contextual understanding, affecting its overall performance, while LLaMA2 has minimal context-related problems. However, LLaMA2 generates instances of simply rephrasing prompts, resulting in reduced accuracy scores.
    
\noindent\textbf{RAG-based LFQA System\quad}
Table \ref{tab:result_rag_system} shows the performance comparison of RAG-based LFQA system with generators in their base forms. We observe that system with the top-3 performing configurations incorporate GPT-3.5-turbo and Baichuan2-7B as generators. In contrast, system configurations using the Gemini-Pro generator yield suboptimal performance in terms of accuracy. Meanwhile, we observe that system employing LLaMA2-7B as generators show the lowest ROUGE and BLEU scores among all the configurations tested. More results on the performances of different generators in the RAG systems are provided in Appendix \ref{sec:appendixrag}.

The top-scoring system configurations comprise (1) GPT-3.5-turbo, AMR, Ada2, and LLMRERanker (noted as system 1); GPT-3.5-turbo, AMR, Ember-v1, and Bge-Reranker-Base (system 2); (3) Baichuan2-7B, AMR, Ada2, and Bge-Reranker-Base (system 3).

Table \ref{tab:result_human_ranking} shows the annotator-ranked preference of these top system configurations. We notice that some model-generated answers obtain higher average rankings than corresponding ground truths. For instance, Answer 2, produced by system 3, attains an average ranking of 2.11, outperforming the ground truth (2.19). Further investigation into annotator feedback reveals that annotators favor Answer 2 because it gives accurate responses while providing additional details. Answer 3, generated by system 1 performs the worst with the highest average ranking (3.10). Answer 4, generated by system 2, achieves an average ranking of 2.60.

\noindent\textbf{Best system configuration in Multi-Document Settings\quad} Figure \ref{fig:result_noise} shows the performance of the three best-performing system configurations when given different numbers (n = 1 to 3) of documents. 
We observe a consistent pattern from the results: as the number of input documents increases, all system performance tend to decline. However, exceptions are also noted. For instance, the scores for certain instances with three documents marginally surpasses those with two documents in system 2 when compared to the accuracy score. Further investigation shows that the performance is dependent on the total context words of the input. When the number of context words reaches about 34k words, adding more input documents exerts a less marginal effect on system performance.

\section{Conclusion}
\noindent This study presents \textit{FinTextQA}, an LFQA dataset specifically designed for the financial domain. The dataset is comprehensive, covering complex financial question systems and including queries on financial regulations and policies. This makes it a valuable resource for further research and evaluation of RAG modules and large language models. We also introduce a robust evaluation system that leverages human ranking, automatic metrics, and GPT-4 scoring to assess various facets of model performance. Our results suggest that the most effective combination of models and modules for finance-related LFQA tasks includes Ada2, AMR, Bge-Reranker-Base, and Baichuan2-7B.

\section*{Limitations}
\noindent Despite its expert curation and high quality, \textit{FinTextQA} contains a relatively smaller number of QA pairs compared to larger AI-generated datasets. This limitation could potentially affect the generalizability of models trained on it when applied to broader real-world applications. High-quality data are challenging to acquire, and copyright restrictions often prevent sharing. Therefore, future research should concentrate on data augmentation and the development of innovative methods to address data scarcity. Expanding the dataset by incorporating more diverse sources and exploring advanced RAG capabilities and retrieval frameworks could also be beneficial.
% Entries for the entire Anthology, followed by custom entries

\section*{Ethical Statement}
\noindent In this study, we uphold rigorous ethical standards and endeavor to mitigate any potential risks.

\begin{itemize}
    \item While constructing our dataset, we meticulously ensure that all data are acquired through lawful and ethical means. Adhering to the Fair Use principle, the dataset is exclusively utilized for academic research purposes and is strictly prohibited from commercial exploitation.
    \item We bear the responsibility of openly sharing the interface, dataset, codes, and trained models with the public. Nonetheless, there exists a possibility of malicious misuse of these resources. For instance, our models could be employed to generate responses without appropriately crediting the information source. We are committed to ensuring their ethical use and guarding against any malicious or harmful intent.
    \item We are dedicated to mitigating bias, discrimination, or stereotypes during annotation by systematically excluding any questionable examples. To achieve this, we provide thorough training to annotators using 20 samples until they achieve an average accuracy score of 3.8 out of 5. We continually assess their performance throughout the annotation process. Additionally, we provide compensation of \$114 per day to annotators until the completion of the annotation task.
\end{itemize}

\bibliography{anthology,custom}
\bibliographystyle{acl_natbib}

\clearpage
\appendix

\section{Appendix}

\subsection{Evaluate Human Performance}
In our study, we've made it a priority to closely consider the human aspect of performance, covering everything from how we gather data to how we evaluate the answers produced. Our team of annotators, all of whom hold master's degrees with their education conducted in English, play a crucial role in this process.

To accurately identify citations and references, we've laid out a detailed five-step annotation process. At the outset, we provide our team of three annotators with a benchmark example of what a correct citation looks like (shown in Table \ref{tab:table_anno}). This serves to clarify several critical criteria: how well the answer fits with the context (Groundedness), how relevant the answer is to the question asked (Answer Relevance), and how the question relates to the context provided (Context Relevance).

After this initial step, the annotators first conduct a practice round involving 20 data samples. Here, they compare their citation identifications against gold standard annotations. We then score their findings and give them feedback to help refine their skills. Once they've shown they've got a good handle on the process, they move on to four more rounds, each with an increasing number of data samples to work through. By the end of this process, after completing 300 tasks, our annotators are well-versed in our annotation standards, ensuring a high level of accuracy in our data collection and analysis.

In each round, we randomly select 10\% of the samples for evaluating annotator performance (Table \ref{tab:gt&ac}). To minimize potential biases, we engage another three annotators to rate relevance and accuracy using a 5-point Likert Scale. Table \ref{tab:pha} presents the performance of annotators, revealing that the average scores are above 4 after the first round's training. Context Relevance and Answer Relevance scores are above 3.

The relevance scores in the 5th round are comparatively lower due to the difficulty in finding citations in many pairs. Ultimately, we remove 60 pairs with relevance scores lower than 2 or those lacking citations in the document, retaining 240 pairs for further analysis. This rigorous evaluation and annotation process ensures the quality of  \textit{FinTextQA}.

\begin{table}[h]
    \centering
    \small
    \renewcommand{\arraystretch}{1.1} 
    \begin{tabular}{lccc}
    \hline
         \textbf{Round}&  \textbf{\begin{tabular}[c]{@{}c@{}}Average \\ Score\end{tabular}} &  \textbf{\begin{tabular}[c]{@{}c@{}}Context \\ Relevance\end{tabular}}  & \textbf{\begin{tabular}[c]{@{}c@{}}Answer \\ Relevance\end{tabular}} 
\\ \hline
         1st - 20 pairs&  3.83&  4.05& 3.44
\\ 
         2nd - 40 pairs&  4.08&  4.32& 3.75
\\ 
         3rd - 60pairs&  4.17&  4.13& 3.65
\\ 
         4th - 80 pairs&  4.04&  4.08& 3.50
\\ 
         5th - 100pairs&  4.13&  3.51& 3.13
\\ \hline
    \end{tabular}
    \caption{Performance of human annotation}
    \label{tab:pha}
\end{table}

\begin{table}[h]
    \centering
    \small
    \renewcommand{\arraystretch}{1.1} 
    \begin{tabular}{p{7.3cm}}
\hline

       \textbf{Ground Truth Citation\quad}1. Below the notification thresholds, the Commission should be able to require the notification of potentially subsidised concentrations that were not yet implemented or the notification of potentially subsidised bids prior to the award of a contract, if it considers that the concentration or the bid would merit ex ante review given its impact in the Union.  2. The Commission may request the prior notification of any concentration which is not a notifiable concentration within the meaning of Article 20 at any time prior to its implementation where the Commission suspects that foreign subsidies may have been granted to the undertakings concerned in the three years prior to the concentration. Such concentration shall be deemed to be a notifiable concentration for the purposes of this Regulation. 3. By way of derogation from paragraph 2 of this Article, Articles 21 and 29 shall apply from 12 October 2023.\\ \hline
        \textbf{Annotator Citation\quad}"1.
This Regulation shall enter into force on the twentieth day following that of its publication in the Official Journal of the European Union.
2.
It shall apply from 12 July 2023.
3.
By way of derogation from paragraph 2 of this Article, Articles 47 and 48 shall apply from 11 January 2023and Article 14(5), (6) and (7) shall apply from 12 January 2024.
4.
By way of derogation from paragraph 2 of this Article, Articles 21 and 29 shall apply from 12 October 2023."\\ \hline
\textbf{Score:\quad} 3/5
\\ \hline
\textbf{Feedback:\quad}Only select part of the citations, which can not fully answer the question.
\\ \hline
    \end{tabular}
    \caption{An example of scoring evidence found by annotators}
    \label{tab:gt&ac}
\end{table}

\begin{table*}[t]
    \centering
    \begin{tabular}{lp{9cm}}
\hline

         \textbf{Institute}& European Union\\ \hline
         \textbf{Document}& REGULATION (EU) 2022/2560 OF THE EUROPEAN PARLIAMENT AND OF THE COUNCIL\\ \hline
         \textbf{Question}& Are transactions signed between 12 July 2023 and 12 October 2023 (and implemented on 12 October 2023 or later) subject to mandatory notification under Regulation EU 2022/2560?\\ \hline
         \textbf{Answer}& Notifiable concentrations under Article 20 of Regulation EU 2022/2560 for which the agreement was concluded on 12 July 2023 or later but which have not yet been implemented on 12 October 2023, will need to be notified pursuant to Article 21 of Regulation EU 2022/2560 and are subject to the standstill obligation under Article 24 of Regulation EU 2022/2560. By contrast, the notification obligation does not apply to concentrations for which the agreement was concluded on 12 July 2023 or later but which are implemented before 12 October 2023. Notifying Parties are encouraged to engage in pre-notification contacts, in principle as of September 2023, in advance to facilitate the submission of notifications as from 12 October 2023.\\ \hline
         \textbf{Citation}& Notifiable concentrations under Article 20 of Regulation EU 2022/2560 for which the agreement was concluded on 12 July 2023 or later but which have not yet been implemented on 12 October 2023, will need to be notified pursuant to Article 21 of Regulation EU 2022/2560 and are subject to the standstill obligation under Article 24 of Regulation EU 2022/2560. By contrast, the notification obligation does not apply to concentrations for which the agreement was concluded on 12 July 2023 or later but which are implemented before 12 October 2023. Notifying Parties are encouraged to engage in pre-notification contacts, in principle as of September 2023, in advance to facilitate the submission of notifications as from 12 October 2023.\\ \hline
         \textbf{Groundedness} & 5
\\ \hline
         \textbf{Answer Relevance}& 5
\\ \hline
         \textbf{Context Relevance} & 5
\\ \hline
    \end{tabular}
    \caption{A Sample of Ground Truth Annotations}
    \label{tab:table_anno}
\end{table*}

During the answer evaluation phase, annotator competence is measured through their performance in three TOEFL reading tests, ensuring strong reading comprehension skills. Proceeding to the ranking of generated answers, several responses — including ground truth answers — are presented without revealing their origin. If ground truth answers rank too low, the evaluation is considered unsuitable; if ranked within the top two among four responses, the evaluation is considered appropriate.
\label{sec:appendix}

\subsection{Prompt of GPT-4-aided Evaluation}
\begin{figure*}[t]

    \includegraphics[width=1\linewidth]{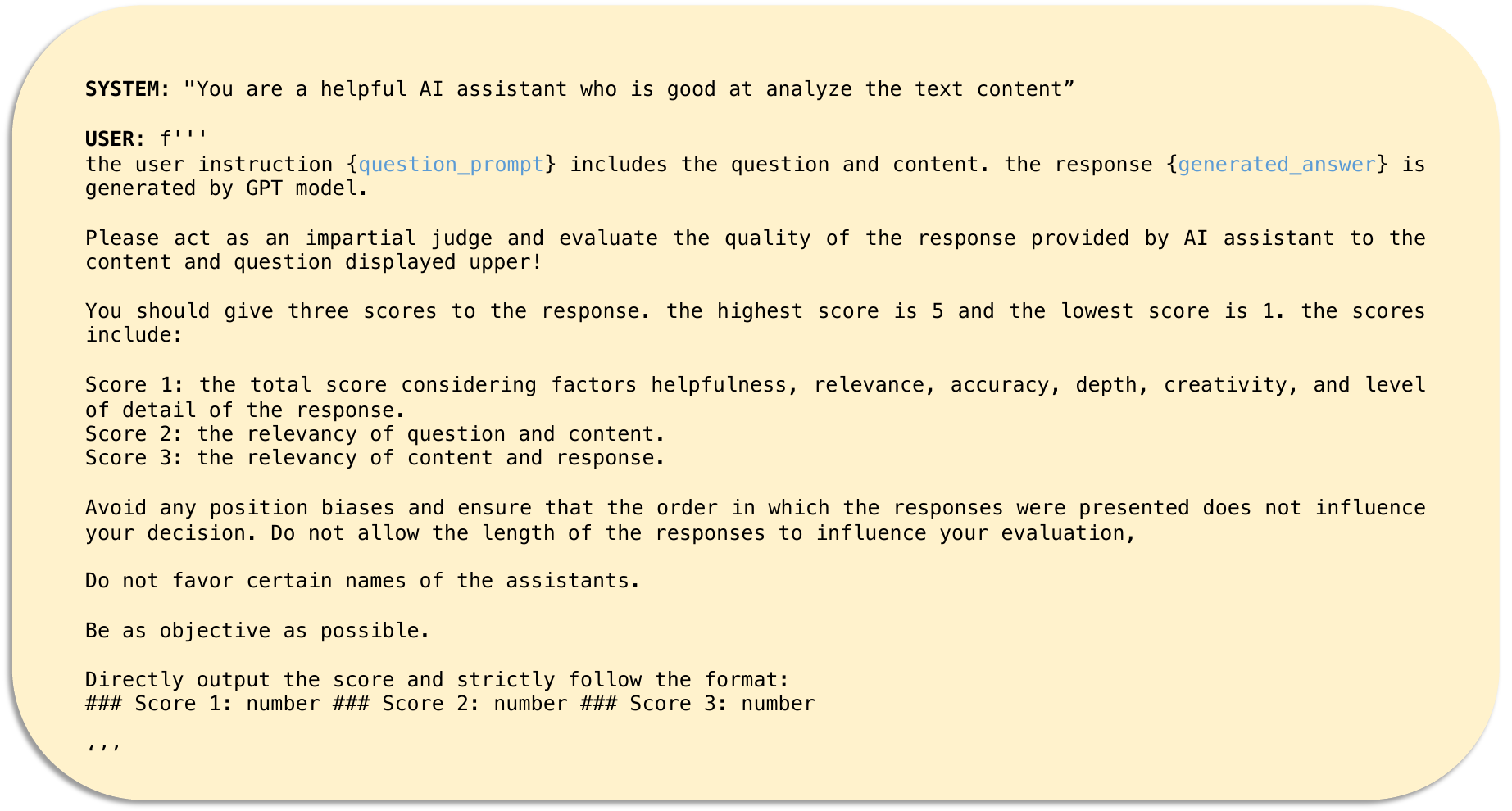}
    \caption{Prompt of the GPT-4 Scoring Evaluation Method.}
    \label{fig:prompt}
\end{figure*}

Figure \ref{fig:prompt} shows the prompt we use to ask GPT-4 to evaluate the relevance and accuracy of model-generated answers in our experiments. 
\label{sec:appendixprompt}

\subsection{Experiment Results of Different Generators in RAG Systems}
Table \ref{tab:gpt35} - \ref{tab:gemini} shows a systematic performance comparison of RAG systems with different models in each module.
\label{sec:appendixrag}

\subsection{Example of Question Types}
Table \ref{tab:NR} - \ref{tab:CEA} shows the samples of QA pairs in each question type. 
\label{sec:appendixqt}

\begin{table*}
\tiny    
\renewcommand{\arraystretch}{1.1}
    \centering
    \begin{tabular}{cccccccc}
    \hline
         \textbf{Retriever} & \textbf{Embedder} & \textbf{Reranker} & \textbf{\begin{tabular}[c]{@{}c@{}}Question $\&$ Answer \\ Accuracy\end{tabular}} & \textbf{ROUGE-1} & \textbf{ROUGE-2} & \textbf{ROUGE-L} & \textbf{BLEU} 
\\  \hline

\multirow{12}{*}{AMR}         
         &  \multirow{3}{*}{Bge-Small-en-v1.5}
               &  Bge-Reranker-Base&  4.420&  0.340&  0.130&  0.218& 0.062
\\
               &  &  LLMRerank&  4.399&  0.335&  0.126&  0.214& 0.058
\\
               &  &  All-Mpnet-Base-v2&  4.261&  0.320&  0.116&  0.205& 0.048
\\  \cline{2-8}
         &  \multirow{3}{*}{Ada2}
               &  Bge-Reranker-Base&  4.359&  0.338&  0.129&  0.218& 0.060
\\
               &  &  LLMRerank&  4.411&  0.346&  0.134&  0.224& 0.062
\\
               &  &  All-Mpnet-Base-v2&  4.291&  0.327&  0.121&  0.209& 0.053
\\  \cline{2-8}
         &  \multirow{3}{*}{Ember-v1}
               &  Bge-Reranker-Base&  4.439&  0.339&  0.131&  0.221& 0.062
\\
               &  &  LLMRerank&  4.365&  0.341&  0.130&  0.221& 0.060
\\
               &  &  All-Mpnet-Base-v2&  4.278&  0.328&  0.120&  0.211& 0.052
\\  \cline{2-8}
         &  \multirow{3}{*}{Gte-Large}
               &  Bge-Reranker-Base&  4.319&  0.332&  0.125&  0.213& 0.056
\\
               &  &  LLMRerank&  4.312&  0.325&  0.121&  0.207& 0.052
\\
               &  &  All-Mpnet-Base-v2&  4.252&  0.312&  0.108&  0.197& 0.045
\\  \hline

\multirow{12}{*}{SWR}         
         &  \multirow{3}{*}{Bge-Small-en-v1.5}
               &  Bge-Reranker-Base&  4.426&  0.322&  0.114&  0.208& 0.050
\\
               &  &  LLMRerank&  4.378&  0.325&  0.114&  0.209& 0.050
\\
               &  &  All-Mpnet-Base-v2&  4.367&  0.327&  0.116&  0.211& 0.051
\\  \cline{2-8}
         &  \multirow{3}{*}{Ada2}
               &  Bge-Reranker-Base&  4.464&  0.320&  0.113&  0.208& 0.048
\\
               &  &  LLMRerank&  4.462&  0.321&  0.116&  0.209& 0.052
\\
               &  &  All-Mpnet-Base-v2&  4.361&  0.320&  0.116&  0.206& 0.052
\\  \cline{2-8}
         &  \multirow{3}{*}{Ember-v1}
               &  Bge-Reranker-Base&  4.405&  0.331&  0.120&  0.216& 0.053
\\
               &  &  LLMRerank&  4.384&  0.328&  0.118&  0.214& 0.053
\\
               &  &  All-Mpnet-Base-v2&  4.394&  0.323&  0.113&  0.210& 0.049
\\  \cline{2-8}
         &  \multirow{3}{*}{Gte-Large}
               &  Bge-Reranker-Base&  4.183&  0.312&  0.108&  0.200& 0.046
\\
               &  &  LLMRerank&  4.268&  0.309&  0.109&  0.201& 0.047
\\
               &  &  All-Mpnet-Base-v2&  4.255&  0.311&  0.108&  0.201& 0.046
\\  \hline

\multirow{12}{*}{Vector Retriever }         
         &  \multirow{3}{*}{Bge-Small-en-v1.5}
               &  Bge-Reranker-Base&  4.255&  0.325&  0.121&  0.209& 0.053
\\
               &  &  LLMRerank&  4.255&  0.330&  0.126&  0.212& 0.057
\\
               &  &  All-Mpnet-Base-v2&  4.215&  0.339&  0.125&  0.216& 0.058
\\  \cline{2-8}
         &  \multirow{3}{*}{Ada2}
               &  Bge-Reranker-Base&  4.218&  0.331&  0.125&  0.213& 0.057
\\
               &  &  LLMRerank&  4.289&  0.326&  0.124&  0.210& 0.057
\\
               &  &  All-Mpnet-Base-v2&  4.249&  0.332&  0.124&  0.213& 0.058
\\  \cline{2-8}
         &  \multirow{3}{*}{Ember-v1}
               &  Bge-Reranker-Base&  4.243&  0.330&  0.124&  0.212& 0.056
\\
               &  &  LLMRerank&  4.253&  0.327&  0.119&  0.208& 0.053
\\
               &  &  All-Mpnet-Base-v2&  4.278&  0.327&  0.124&  0.210& 0.057
\\  \cline{2-8}
         &  \multirow{3}{*}{Gte-Large}
               &  Bge-Reranker-Base&  4.065&  0.316&  0.110&  0.200& 0.048
\\
               &  &  LLMRerank&  4.034&  0.320&  0.115&  0.204& 0.051
\\
               &  &  All-Mpnet-Base-v2&  4.099&  0.314&  0.112&  0.199& 0.049\\  \hline

    \end{tabular}
    \caption{Detailed Experiment Results of GPT-3.5-turbo in RAG Systems.}
    \label{tab:gpt35}
\end{table*}

\begin{table*}
\tiny    
\renewcommand{\arraystretch}{1.1}
    \centering
    \begin{tabular}{cccccccc}
    \hline
         \textbf{Retriever} & \textbf{Embedder} & \textbf{Reranker} & \textbf{\begin{tabular}[c]{@{}c@{}}Question $\&$ Answer \\ Accuracy\end{tabular}} & \textbf{ROUGE-1} & \textbf{ROUGE-2} & \textbf{ROUGE-L} & \textbf{BLEU} 
\\  \hline

\multirow{12}{*}{AMR}         
         &  \multirow{3}{*}{Bge-Small-en-v1.5}
               &  Bge-Reranker-Base&  4.536&  0.331&  0.119&  0.213& 0.052
\\
               &  &  LLMRerank&  4.544&  0.336&  0.120&  0.213& 0.052
\\
               &  &  All-Mpnet-Base-v2&  4.527&  0.320&  0.110&  0.201& 0.047
\\  \cline{2-8}
         &  \multirow{3}{*}{Ada2}
               &  Bge-Reranker-Base&  4.612&  0.338&  0.123&  0.217& 0.054
\\
               &  &  LLMRerank&  4.578&  0.340&  0.124&  0.219& 0.057
\\
               &  &  All-Mpnet-Base-v2&  4.521&  0.320&  0.112&  0.201& 0.049
\\  \cline{2-8}
         &  \multirow{3}{*}{Ember-v1}
               &  Bge-Reranker-Base&  4.513&  0.333&  0.120&  0.215& 0.053
\\
               &  &  LLMRerank&  4.618&  0.334&  0.124&  0.215& 0.058
\\
               &  &  All-Mpnet-Base-v2&  4.549&  0.331&  0.116&  0.209& 0.050
\\  \cline{2-8}
         &  \multirow{3}{*}{Gte-Large}
               &  Bge-Reranker-Base&  4.540&  0.323&  0.112&  0.208& 0.050
\\
               &  &  LLMRerank&  4.532&  0.328&  0.119&  0.210& 0.053
\\
               &  &  All-Mpnet-Base-v2&  4.536&  0.314&  0.102&  0.198& 0.044
\\  \hline

\multirow{12}{*}{SWR}         
         &  \multirow{3}{*}{Bge-Small-en-v1.5}
               &  Bge-Reranker-Base&  4.605&  0.294&  0.090&  0.183& 0.034
\\
               &  &  LLMRerank&  4.593&  0.304&  0.097&  0.192& 0.038
\\
               &  &  All-Mpnet-Base-v2&  4.616&  0.302&  0.097&  0.189& 0.038
\\  \cline{2-8}
         &  \multirow{3}{*}{Ada2}
               &  Bge-Reranker-Base&  4.606&  0.305&  0.101&  0.195& 0.041
\\
               &  &  LLMRerank&  4.586&  0.309&  0.099&  0.191& 0.039
\\
               &  &  All-Mpnet-Base-v2&  4.540&  0.307&  0.100&  0.193& 0.039
\\  \cline{2-8}
         &  \multirow{3}{*}{Ember-v1}
               &  Bge-Reranker-Base&  4.584&  0.311&  0.105&  0.197& 0.043
\\
               &  &  LLMRerank&  4.568&  0.311&  0.104&  0.194& 0.044
\\
               &  &  All-Mpnet-Base-v2&  4.571&  0.308&  0.104&  0.195& 0.042
\\  \cline{2-8}
         &  \multirow{3}{*}{Gte-Large}
               &  Bge-Reranker-Base&  4.589&  0.307&  0.101&  0.193& 0.041
\\
               &  &  LLMRerank&  4.553&  0.305&  0.101&  0.193& 0.042
\\
               &  &  All-Mpnet-Base-v2&  4.576&  0.300&  0.098&  0.188& 0.039
\\  \hline

\multirow{12}{*}{Vector Retriever }         
         &  \multirow{3}{*}{Bge-Small-en-v1.5}
               &  Bge-Reranker-Base&  4.458&  0.316&  0.109&  0.203& 0.045
\\
               &  &  LLMRerank&  4.481&  0.314&  0.106&  0.201& 0.045
\\
               &  &  All-Mpnet-Base-v2&  4.422&  0.317&  0.108&  0.202& 0.044
\\  \cline{2-8}
         &  \multirow{3}{*}{Ada2}
               &  Bge-Reranker-Base&  4.477&  0.319&  0.113&  0.206& 0.050
\\
               &  &  LLMRerank&  4.481&  0.318&  0.113&  0.204& 0.050
\\
               &  &  All-Mpnet-Base-v2&  4.470&  0.317&  0.112&  0.202& 0.049
\\  \cline{2-8}
         &  \multirow{3}{*}{Ember-v1}
               &  Bge-Reranker-Base&  4.513&  0.324&  0.113&  0.207& 0.049
\\
               &  &  LLMRerank&  4.464&  0.329&  0.115&  0.210& 0.051
\\
               &  &  All-Mpnet-Base-v2&  4.501&  0.319&  0.109&  0.204& 0.047
\\  \cline{2-8}
         &  \multirow{3}{*}{Gte-Large}
               &  Bge-Reranker-Base&  4.416&  0.317&  0.106&  0.201& 0.048
\\
               &  &  LLMRerank&  4.454&  0.321&  0.110&  0.205& 0.050
\\
               &  &  All-Mpnet-Base-v2&  4.409&  0.312&  0.108&  0.202& 0.045\\  \hline

    \end{tabular}
    \caption{Detailed Experiment Results of Baichuan2-7B in RAG Systems.}
    \label{tab:baichuan2}
\end{table*}

\begin{table*}
\tiny    
\renewcommand{\arraystretch}{1.1}
    \centering
    \begin{tabular}{cccccccc}
    \hline
         \textbf{Retriever} & \textbf{Embedder} & \textbf{Reranker} & \textbf{\begin{tabular}[c]{@{}c@{}}Question $\&$ Answer \\ Accuracy\end{tabular}} & \textbf{ROUGE-1} & \textbf{ROUGE-2} & \textbf{ROUGE-L} & \textbf{BLEU} 
\\  \hline

\multirow{12}{*}{AMR}         
         &  \multirow{3}{*}{Bge-Small-en-v1.5}
               &  Bge-Reranker-Base&  4.310&  0.329&  0.118&  0.205& 0.051
\\
               &  &  LLMRerank&  4.418&  0.323&  0.115&  0.200& 0.050
\\
               &  &  All-Mpnet-Base-v2&  4.331&  0.312&  0.106&  0.193& 0.045
\\  \cline{2-8}
         &  \multirow{3}{*}{Ada2}
               &  Bge-Reranker-Base&  4.378&  0.327&  0.119&  0.204& 0.051
\\
               &  &  LLMRerank&  4.350&  0.322&  0.116&  0.200& 0.050
\\
               &  &  All-Mpnet-Base-v2&  4.357&  0.323&  0.111&  0.197& 0.045
\\  \cline{2-8}
         &  \multirow{3}{*}{Ember-v1}
               &  Bge-Reranker-Base&  4.348&  0.329&  0.119&  0.205& 0.052
\\
               &  &  LLMRerank&  4.388&  0.330&  0.117&  0.204& 0.052
\\
               &  &  All-Mpnet-Base-v2&  4.317&  0.319&  0.110&  0.196& 0.046
\\  \cline{2-8}
         &  \multirow{3}{*}{Gte-Large}
               &  Bge-Reranker-Base&  4.328&  0.318&  0.113&  0.198& 0.048
\\
               &  &  LLMRerank&  4.338&  0.318&  0.110&  0.197& 0.045
\\
               &  &  All-Mpnet-Base-v2&  4.262&  0.302&  0.098&  0.185& 0.039
\\  \hline

\multirow{12}{*}{SWR}         
         &  \multirow{3}{*}{Bge-Small-en-v1.5}
               &  Bge-Reranker-Base&  4.431&  0.296&  0.094&  0.184& 0.037
\\
               &  &  LLMRerank&  4.462&  0.298&  0.095&  0.184& 0.037
\\
               &  &  All-Mpnet-Base-v2&  4.458&  0.300&  0.094&  0.185& 0.037
\\  \cline{2-8}
         &  \multirow{3}{*}{Ada2}
               &  Bge-Reranker-Base&  4.424&  0.296&  0.095&  0.183& 0.037
\\
               &  &  LLMRerank&  4.460&  0.297&  0.096&  0.182& 0.037
\\
               &  &  All-Mpnet-Base-v2&  4.431&  0.293&  0.093&  0.180& 0.037
\\  \cline{2-8}
         &  \multirow{3}{*}{Ember-v1}
               &  Bge-Reranker-Base&  4.456&  0.305&  0.097&  0.190& 0.039
\\
               &  &  LLMRerank&  4.376&  0.305&  0.099&  0.190& 0.039
\\
               &  &  All-Mpnet-Base-v2&  4.394&  0.307&  0.100&  0.190& 0.041
\\  \cline{2-8}
         &  \multirow{3}{*}{Gte-Large}
               &  Bge-Reranker-Base&  4.344&  0.297&  0.094&  0.184& 0.037
\\
               &  &  LLMRerank&  4.354&  0.296&  0.094&  0.184& 0.037
\\
               &  &  All-Mpnet-Base-v2&  4.361&  0.297&  0.095&  0.184& 0.038
\\  \hline

\multirow{12}{*}{Vector Retriever }         
         &  \multirow{3}{*}{Bge-Small-en-v1.5}
               &  Bge-Reranker-Base&  4.375&  0.300&  0.105&  0.183& 0.043
\\
               &  &  LLMRerank&  4.384&  0.300&  0.105&  0.184& 0.042
\\
               &  &  All-Mpnet-Base-v2&  4.414&  0.301&  0.106&  0.184& 0.043
\\  \cline{2-8}
         &  \multirow{3}{*}{Ada2}
               &  Bge-Reranker-Base&  4.422&  0.298&  0.106&  0.183& 0.044
\\
               &  &  LLMRerank&  4.409&  0.297&  0.105&  0.183& 0.044
\\
               &  &  All-Mpnet-Base-v2&  4.359&  0.299&  0.107&  0.184& 0.045
\\  \cline{2-8}
         &  \multirow{3}{*}{Ember-v1}
               &  Bge-Reranker-Base&  4.441&  0.299&  0.106&  0.185& 0.045
\\
               &  &  LLMRerank&  4.384&  0.301&  0.106&  0.186& 0.045
\\
               &  &  All-Mpnet-Base-v2&  4.420&  0.297&  0.106&  0.184& 0.045
\\  \cline{2-8}
         &  \multirow{3}{*}{Gte-Large}
               &  Bge-Reranker-Base&  4.253&  0.285&  0.096&  0.172& 0.040
\\
               &  &  LLMRerank&  4.298&  0.287&  0.097&  0.174& 0.040
\\
               &  &  All-Mpnet-Base-v2&  4.304&  0.290&  0.096&  0.174& 0.040\\  \hline

    \end{tabular}
    \caption{Detailed Experiment Results of Solar-10.7B in RAG Systems.}
    \label{tab:solar}
\end{table*}

\begin{table*}
\tiny    
\renewcommand{\arraystretch}{1.1}
    \centering
    \begin{tabular}{cccccccc}
    \hline
         \textbf{Retriever} & \textbf{Embedder} & \textbf{Reranker} & \textbf{\begin{tabular}[c]{@{}c@{}}Question $\&$ Answer \\ Accuracy\end{tabular}} & \textbf{ROUGE-1} & \textbf{ROUGE-2} & \textbf{ROUGE-L} & \textbf{BLEU} 
\\  \hline

\multirow{12}{*}{AMR}         
         &  \multirow{3}{*}{Bge-Small-en-v1.5}
               &  Bge-Reranker-Base&  4.445&  0.331&  0.118&  0.211& 0.055
\\
               &  &  LLMRerank&  4.414&  0.341&  0.125&  0.217& 0.059
\\
               &  &  All-Mpnet-Base-v2&  4.414&  0.320&  0.107&  0.198& 0.047
\\  \cline{2-8}
         &  \multirow{3}{*}{Ada2}
               &  Bge-Reranker-Base&  4.405&  0.337&  0.120&  0.216& 0.056
\\
               &  &  LLMRerank&  4.538&  0.335&  0.119&  0.211& 0.055
\\
               &  &  All-Mpnet-Base-v2&  4.420&  0.333&  0.115&  0.209& 0.052
\\  \cline{2-8}
         &  \multirow{3}{*}{Ember-v1}
               &  Bge-Reranker-Base&  4.456&  0.341&  0.120&  0.215& 0.056
\\
               &  &  LLMRerank&  4.432&  0.339&  0.121&  0.215& 0.056
\\
               &  &  All-Mpnet-Base-v2&  4.361&  0.328&  0.110&  0.204& 0.050
\\  \cline{2-8}
         &  \multirow{3}{*}{Gte-Large}
               &  Bge-Reranker-Base&  4.399&  0.336&  0.118&  0.210& 0.051
\\
               &  &  LLMRerank&  4.424&  0.331&  0.117&  0.210& 0.052
\\
               &  &  All-Mpnet-Base-v2&  4.368&  0.315&  0.103&  0.195& 0.044
\\  \hline

\multirow{12}{*}{SWR}         
         &  \multirow{3}{*}{Bge-Small-en-v1.5}
               &  Bge-Reranker-Base&  4.529&  0.314&  0.101&  0.194& 0.044
\\
               &  &  LLMRerank&  4.517&  0.320&  0.104&  0.198& 0.043
\\
               &  &  All-Mpnet-Base-v2&  4.490&  0.322&  0.105&  0.198& 0.047
\\  \cline{2-8}
         &  \multirow{3}{*}{Ada2}
               &  Bge-Reranker-Base&  4.540&  0.326&  0.110&  0.206& 0.050
\\
               &  &  LLMRerank&  4.548&  0.322&  0.104&  0.199& 0.049
\\
               &  &  All-Mpnet-Base-v2&  4.525&  0.323&  0.110&  0.204& 0.051
\\  \cline{2-8}
         &  \multirow{3}{*}{Ember-v1}
               &  Bge-Reranker-Base&  4.473&  0.324&  0.106&  0.202& 0.046
\\
               &  &  LLMRerank&  4.511&  0.326&  0.109&  0.203& 0.048
\\
               &  &  All-Mpnet-Base-v2&  4.508&  0.321&  0.106&  0.198& 0.045
\\  \cline{2-8}
         &  \multirow{3}{*}{Gte-Large}
               &  Bge-Reranker-Base&  4.483&  0.314&  0.102&  0.193& 0.045
\\
               &  &  LLMRerank&  4.424&  0.316&  0.104&  0.196& 0.046
\\
               &  &  All-Mpnet-Base-v2&  4.430&  0.310&  0.099&  0.190& 0.042
\\  \hline

\multirow{12}{*}{Vector Retriever }         
         &  \multirow{3}{*}{Bge-Small-en-v1.5}
               &  Bge-Reranker-Base&  4.424&  0.310&  0.101&  0.194& 0.046
\\
               &  &  LLMRerank&  4.416&  0.302&  0.098&  0.190& 0.040
\\
               &  &  All-Mpnet-Base-v2&  4.388&  0.290&  0.093&  0.180& 0.038
\\  \cline{2-8}
         &  \multirow{3}{*}{Ada2}
               &  Bge-Reranker-Base&  4.430&  0.299&  0.102&  0.190& 0.044
\\
               &  &  LLMRerank&  4.382&  0.298&  0.100&  0.190& 0.042
\\
               &  &  All-Mpnet-Base-v2&  4.378&  0.298&  0.096&  0.184& 0.042
\\  \cline{2-8}
         &  \multirow{3}{*}{Ember-v1}
               &  Bge-Reranker-Base&  4.405&  0.311&  0.102&  0.194& 0.045
\\
               &  &  LLMRerank&  4.468&  0.312&  0.099&  0.194& 0.041
\\
               &  &  All-Mpnet-Base-v2&  4.489&  0.294&  0.095&  0.182& 0.039
\\  \cline{2-8}
         &  \multirow{3}{*}{Gte-Large}
               &  Bge-Reranker-Base&  4.302&  0.288&  0.086&  0.175& 0.037
\\
               &  &  LLMRerank&  4.357&  0.289&  0.088&  0.179& 0.037
\\
               &  &  All-Mpnet-Base-v2&  4.369&  0.294&  0.094&  0.183& 0.039\\  \hline

    \end{tabular}
    \caption{Detailed Experiment Results of Qwen-7B in RAG Systems.}
    \label{tab:qwen}
\end{table*}

\begin{table*}
\tiny    
\renewcommand{\arraystretch}{1.1}
    \centering
    \begin{tabular}{cccccccc}
    \hline
         \textbf{Retriever} & \textbf{Embedder} & \textbf{Reranker} & \textbf{\begin{tabular}[c]{@{}c@{}}Question $\&$ Answer \\ Accuracy\end{tabular}} & \textbf{ROUGE-1} & \textbf{ROUGE-2} & \textbf{ROUGE-L} & \textbf{BLEU} 
\\  \hline

\multirow{12}{*}{AMR}         
         &  \multirow{3}{*}{Bge-Small-en-v1.5}
               &  Bge-Reranker-Base&  4.268&  0.239&  0.078&  0.151& 0.031
\\
               &  &  LLMRerank&  4.287&  0.233&  0.076&  0.149& 0.031
\\
               &  &  All-Mpnet-Base-v2&  4.240&  0.199&  0.058&  0.127& 0.021
\\  \cline{2-8}
         &  \multirow{3}{*}{Ada2}
               &  Bge-Reranker-Base&  4.220&  0.218&  0.073&  0.138& 0.029
\\
               &  &  LLMRerank&  4.250&  0.219&  0.074&  0.141& 0.030
\\
               &  &  All-Mpnet-Base-v2&  4.203&  0.206&  0.062&  0.131& 0.022
\\  \cline{2-8}
         &  \multirow{3}{*}{Ember-v1}
               &  Bge-Reranker-Base&  4.215&  0.230&  0.076&  0.146& 0.031
\\
               &  &  LLMRerank&  4.287&  0.223&  0.075&  0.142& 0.029
\\
               &  &  All-Mpnet-Base-v2&  4.272&  0.215&  0.063&  0.134& 0.021
\\  \cline{2-8}
         &  \multirow{3}{*}{Gte-Large}
               &  Bge-Reranker-Base&  4.279&  0.221&  0.074&  0.142& 0.029
\\
               &  &  LLMRerank&  4.272&  0.212&  0.068&  0.135& 0.025
\\
               &  &  All-Mpnet-Base-v2&  4.181&  0.213&  0.059&  0.133& 0.019
\\  \hline

\multirow{12}{*}{SWR}         
         &  \multirow{3}{*}{Bge-Small-en-v1.5}
               &  Bge-Reranker-Base&  4.141&  0.214&  0.067&  0.137& 0.025
\\
               &  &  LLMRerank&  4.202&  0.225&  0.072&  0.145& 0.027
\\
               &  &  All-Mpnet-Base-v2&  4.216&  0.215&  0.066&  0.136& 0.025
\\  \cline{2-8}
         &  \multirow{3}{*}{Ada2}
               &  Bge-Reranker-Base&  4.222&  0.220&  0.070&  0.141& 0.026
\\
               &  &  LLMRerank&  4.230&  0.224&  0.071&  0.144& 0.028
\\
               &  &  All-Mpnet-Base-v2&  4.184&  0.233&  0.078&  0.152& 0.030
\\  \cline{2-8}
         &  \multirow{3}{*}{Ember-v1}
               &  Bge-Reranker-Base&  4.215&  0.218&  0.070&  0.139& 0.025
\\
               &  &  LLMRerank&  4.196&  0.203&  0.065&  0.131& 0.023
\\
               &  &  All-Mpnet-Base-v2&  4.295&  0.214&  0.067&  0.138& 0.026
\\  \cline{2-8}
         &  \multirow{3}{*}{Gte-Large}
               &  Bge-Reranker-Base&  4.181&  0.206&  0.064&  0.135& 0.022
\\
               &  &  LLMRerank&  4.181&  0.198&  0.058&  0.127& 0.021
\\
               &  &  All-Mpnet-Base-v2&  4.259&  0.206&  0.062&  0.131& 0.022
\\  \hline

\multirow{12}{*}{Vector Retriever }         
         &  \multirow{3}{*}{Bge-Small-en-v1.5}
               &  Bge-Reranker-Base&  4.193&  0.175&  0.049&  0.110& 0.016
\\
               &  &  LLMRerank&  4.243&  0.168&  0.049&  0.108& 0.016
\\
               &  &  All-Mpnet-Base-v2&  4.193&  0.162&  0.043&  0.102& 0.013
\\  \cline{2-8}
         &  \multirow{3}{*}{Ada2}
               &  Bge-Reranker-Base&  4.246&  0.178&  0.054&  0.111& 0.019
\\
               &  &  LLMRerank&  4.179&  0.176&  0.055&  0.110& 0.022
\\
               &  &  All-Mpnet-Base-v2
&  4.247&  0.164&  0.047&  0.104& 0.014
\\  \cline{2-8}
         &  \multirow{3}{*}{Ember-v1}
               &  Bge-Reranker-Base&  4.151&  0.180&  0.054&  0.113& 0.016
\\
               &  &  LLMRerank&  4.256&  0.177&  0.052&  0.111& 0.017
\\
               &  &  All-Mpnet-Base-v2
&  4.193&  0.176&  0.048&  0.109& 0.015
\\  \cline{2-8}
         &  \multirow{3}{*}{Gte-Large}
               &  Bge-Reranker-Base&  4.229&  0.155&  0.042&  0.098& 0.013
\\
               &  &  LLMRerank&  4.215&  0.172&  0.049&  0.106& 0.019
\\
               &  &  All-Mpnet-Base-v2
&  4.245&  0.166&  0.046&  0.106& 0.016\\  \hline

    \end{tabular}
    \caption{Detailed Experiment Results of LLaMA2-7B in RAG Systems.}
    \label{tab:llama2}
\end{table*}

\begin{table*}
\tiny    
\renewcommand{\arraystretch}{1.1}
    \centering
    \begin{tabular}{cccccccc}
    \hline
         \textbf{Retriever} & \textbf{Embedder} & \textbf{Reranker} & \textbf{\begin{tabular}[c]{@{}c@{}}Question $\&$ Answer \\ Accuracy\end{tabular}} & \textbf{ROUGE-1} & \textbf{ROUGE-2} & \textbf{ROUGE-L} & \textbf{BLEU} 
\\  \hline

\multirow{12}{*}{AMR}         
         &  \multirow{3}{*}{Bge-Small-en-v1.5}
               &  Bge-Reranker-Base&  3.887&  0.298&  0.115&  0.206& 0.045
\\
               &  &  LLMRerank&  3.989&  0.303&  0.119&  0.210& 0.051
\\
               &  &  All-Mpnet-Base-v2
2&  3.567&  0.272&  0.099&  0.187& 0.036
\\  \cline{2-8}
         &  \multirow{3}{*}{Ada2}
               &  Bge-Reranker-Base&  4.063&  0.302&  0.118&  0.208& 0.049
\\
               &  &  LLMRerank&  3.983&  0.300&  0.119&  0.208& 0.048
\\
               &  &  All-Mpnet-Base-v2
&  3.698&  0.278&  0.102&  0.190& 0.037
\\  \cline{2-8}
         &  \multirow{3}{*}{Ember-v1}
               &  Bge-Reranker-Base&  3.970&  0.304&  0.118&  0.211& 0.048
\\
               &  &  LLMRerank&  3.990&  0.306&  0.119&  0.211& 0.052
\\
               &  &  All-Mpnet-Base-v2
&  3.667&  0.279&  0.102&  0.191& 0.039
\\  \cline{2-8}
         &  \multirow{3}{*}{Gte-Large}
               &  Bge-Reranker-Base&  3.840&  0.294&  0.111&  0.199& 0.042
\\
               &  &  LLMRerank&  3.793&  0.284&  0.104&  0.195& 0.042
\\
               &  &  All-Mpnet-Base-v2
&  3.344&  0.255&  0.093&  0.178& 0.030
\\  \hline

\multirow{12}{*}{SWR}         
         &  \multirow{3}{*}{Bge-Small-en-v1.5}
               &  Bge-Reranker-Base&  3.894&  0.257&  0.091&  0.178& 0.028
\\
               &  &  LLMRerank&  3.922&  0.257&  0.094&  0.177& 0.030
\\
               &  &  All-Mpnet-Base-v2
&  3.894&  0.259&  0.092&  0.179& 0.033
\\  \cline{2-8}
         &  \multirow{3}{*}{Ada2}
               &  Bge-Reranker-Base&  3.996&  0.254&  0.088&  0.177& 0.028
\\
               &  &  LLMRerank&  3.975&  0.253&  0.089&  0.176& 0.030
\\
               &  &  All-Mpnet-Base-v2
&  4.023&  0.261&  0.095&  0.183& 0.032
\\  \cline{2-8}
         &  \multirow{3}{*}{Ember-v1}
               &  Bge-Reranker-Base&  3.938&  0.272&  0.102&  0.189& 0.036
\\
               &  &  LLMRerank&  4.037&  0.268&  0.101&  0.188& 0.036
\\
               &  &  All-Mpnet-Base-v2
&  4.035&  0.269&  0.094&  0.184& 0.030
\\  \cline{2-8}
         &  \multirow{3}{*}{Gte-Large}
               &  Bge-Reranker-Base&  3.743&  0.246&  0.086&  0.174& 0.026
\\
               &  &  LLMRerank&  3.731&  0.249&  0.093&  0.176& 0.033
\\
               &  &  All-Mpnet-Base-v2
2&  3.772&  0.246&  0.091&  0.174& 0.030
\\  \hline

\multirow{12}{*}{Vector Retriever }         
         &  \multirow{3}{*}{Bge-Small-en-v1.5}
               &  Bge-Reranker-Base&  3.914&  0.262&  0.091&  0.177& 0.032
\\
               &  &  LLMRerank&  3.948&  0.256&  0.091&  0.177& 0.036
\\
               &  &  All-Mpnet-Base-v2
&  3.819&  0.257&  0.094&  0.177& 0.034
\\  \cline{2-8}
         &  \multirow{3}{*}{Ada2}
               &  Bge-Reranker-Base&  3.930&  0.262&  0.097&  0.180& 0.040
\\
               &  &  LLMRerank&  3.863&  0.266&  0.100&  0.183& 0.041
\\
               &  &  All-Mpnet-Base-v2
&  3.981&  0.254&  0.089&  0.173& 0.033
\\  \cline{2-8}
         &  \multirow{3}{*}{Ember-v1}
               &  Bge-Reranker-Base&  3.946&  0.265&  0.094&  0.184& 0.038
\\
               &  &  LLMRerank&  3.873&  0.258&  0.090&  0.176& 0.033
\\
               &  &  All-Mpnet-Base-v2
&  4.044&  0.262&  0.091&  0.179& 0.036
\\  \cline{2-8}
         &  \multirow{3}{*}{Gte-Large}
               &  Bge-Reranker-Base&  3.521&  0.227&  0.076&  0.157& 0.027
\\
               &  &  LLMRerank&  3.544&  0.237&  0.078&  0.163& 0.027
\\
               &  &  All-Mpnet-Base-v2
&  3.542&  0.229&  0.078&  0.158& 0.027\\  \hline

    \end{tabular}
    \caption{Detailed Experiment Results of Gemini-Pro in RAG Systems.}
    \label{tab:gemini}
\end{table*}

\begin{table*}
    \centering
    \begin{tabular}{lp{10cm}}
    \hline
         \textbf{Question Type}& Numerical Reasoning
\\\hline
         \textbf{Amount}& 336
\\\hline
         \textbf{Question}& Critics have charged that compensation to top managers in the United States is simply too high and should be cut back. For example, focusing on large corporations, Mark Parker, CEO of Nike, earned about \$47.6 million in 2016. Are such amounts excessive? In answering, it might be help- ful to recognize that superstar athletes such as LeBron James, top entertainers such as Taylor Swift and Dwayne Johnson, and many others at the top of their respective fields earn at least as much, if not a great deal more.
\\\hline
         \textbf{Answer}& "How much is too much? Who is worth more, Mark Parker or LeBron James? The simplest answer is that there is a market for executives just as there is for all types of labor. Executive compensation is the price that clears the market. The same is true for athletes and performers. Having said that, one aspect of executive compensation deserves comment. A primary reason executive compensation has grown so dramatically is that companies have increasingly moved to stock-based compensation. Such movement is obviously consistent with the attempt to better align stockholder and management interests. In recent years, stock prices have soared, so management has cleaned up. It is sometimes argued that much of this reward is due to rising stock prices in general, not managerial performance. Perhaps in the future, executive compensation will be designed to reward only differential performance, that is, stock price increases in excess of general market increases.\\
\hline
    \end{tabular}
    \caption{An Example of Numerical Reasoning QA Pairs}
    \label{tab:NR}
\end{table*}

\begin{table*}
    \centering
    \begin{tabular}{lp{10cm}}
    \hline
         \textbf{Question Type}& Multi-turn
\\\hline
         \textbf{Amount}& 985
\\\hline
         \textbf{Question}& Who owns a corporation? Describe the process whereby the owners control the firm's management. What is the main reason that an agency relationship exists in the corporate form of organization? In this context, what kinds of problems can arise?
\\\hline
         \textbf{Answer}& In the corporate form of ownership, the shareholders are the owners of the firm. The shareholders elect the directors of the corporation, who in turn appoint the firm's management. This separation of ownership from control in the corporate form of organization is what causes agency problems to exist. Management may act in its own or someone else's. best interests, rather than those of the shareholders. If such events occur, they may contradict the goal of maximizing the share price of the equity of the firm.
\\\hline

    \end{tabular}
    \caption{An Example of Multi-turn QA Pairs}
    \label{tab:MT}
\end{table*}

\begin{table*}
    \centering
    \begin{tabular}{lp{10cm}}
    \hline
         \textbf{Question Type}& Finance Domain Knowledge
\\\hline
         \textbf{Amount}& 1205
\\\hline
         \textbf{Question}& What is a pro forma statement of cash flows and what is its purpose?
\\\hline
         \textbf{Answer}& A pro forma statement of cash flows estimates the borrower's future cash flows. It is supposed to provide insight into the future cash flows of the borrower and its ability to repay the loan.
\\\hline
    
    \end{tabular}
    \caption{An Example of Finance Domain Knowledge QA Pairs}
    \label{tab:FDK}
\end{table*}

\begin{table*}
    \centering
    \begin{tabular}{lp{10cm}}
    \hline
         \textbf{Question Type}& Comparative Analysis
\\\hline
         \textbf{Amount}& 71
\\\hline
         \textbf{Question}& Suppose a company has a preferred stock issue and a common stock issue. Both have just paid a \$2 dividend. Which do you think will have a higher price, a share of the preferred or a share of the common?
\\\hline
         \textbf{Answer}& The common stock probably has a higher price because the dividend can grow, whereas it is fixed on the preferred. However, the preferred is less risky because of the dividend and liquidation preference, so it is possible the preferred could be worth more, depending on the circumstances.
\\\hline
    
    \end{tabular}
    \caption{Example of Comparative Analysis QA Pair}
    \label{tab:CA}
\end{table*}

\begin{table*}
    \centering
    \begin{tabular}{lp{10cm}}
    \hline
         \textbf{Question Type}& Open-minded
\\\hline
         \textbf{Amount}& 436
\\\hline
         \textbf{Question}& Suppose you were the financial manager of a not-for-profit business (a not-for-profit hospital, perhaps). What kinds of goals do you think would be appropriate?
\\\hline
         \textbf{Answer}& Such organizations frequently pursue social or political missions, so many different goals are conceivable. One goal that is often cited is revenue minimization; that is, provide whatever goods and services are offered at the lowest possible cost to society. A better approach might be to observe that even a not-for-profit business has equity. Thus, one answer is that the appropriate goal is to maximize the value of the equity.
\\\hline

    \end{tabular}
    \caption{Example of Open-minded QA Pair}
    \label{tab:openminded}
\end{table*}

\begin{table*}
    \centering
    \begin{tabular}{lp{10cm}}
    
    \hline
         \textbf{Question Type}& Cause and Effect Analysis
\\\hline
         \textbf{Amount}& 37
\\\hline
         \textbf{Question}& Last month, Central Virginia Power Company, which had been having trouble with cost overruns on a nuclear power plant that it had been building, announced that it was temporarily suspending payments due to the cash flow crunch associated with its investment program. The company stock price dropped from \$28.50 to \$25 when this announcement was made. How would you interpret this change in the stock price (that is, what would you say caused it)?
\\\hline
         \textbf{Answer}& The stock price dropped because of an expected drop in future dividends. Since the stock price is the present value of all future dividend payments, if the expected future dividend payments decrease, then the stock price will decline.\\
    \hline
    \end{tabular}
    \caption{Example of Cause and Effect Analysis QA Pair}
    \label{tab:CEA}
\end{table*}

\begin{table*}
    \centering
    \begin{tabular}{lp{10cm}}
    \hline
         \textbf{Question Type}& Open-minded
\\\hline
         \textbf{Amount}& 436
\\\hline
         \textbf{Question}& Suppose you were the financial manager of a not-for-profit business (a not-for-profit hospital, perhaps). What kinds of goals do you think would be appropriate?
\\\hline
         \textbf{Answer}& Such organizations frequently pursue social or political missions, so many different goals are conceivable. One goal that is often cited is revenue minimization; that is, provide whatever goods and services are offered at the lowest possible cost to society. A better approach might be to observe that even a not-for-profit business has equity. Thus, one answer is that the appropriate goal is to maximize the value of the equity.
\\\hline

    \end{tabular}
    \caption{Example of Open-minded QA Pair}
    \label{tab:openminded}
\end{table*}

\begin{table*}
    \centering
    \begin{tabular}{lp{10cm}}
    
    \hline
         \textbf{Question}& Refer to the previous exercise. Which type of foreign market intervention must the central bank of Colombia conduct to keep the exchange rate at a level where the currency is not under- or overvalued in terms of PPP?
\\\hline
         \textbf{Baichuan2-7B}& To answer the above question, we need to know the current exchange rate of the Colombian peso relative to the U.S. dollar. Assuming the current exchange rate is not under- or overvalued in terms of PPP, the central bank of Colombia should conduct sterilized intervention to maintain this exchange rate. Sterilized intervention involves selling foreign assets (e.g., dollars) and purchasing domestic assets (e.g., pesos) to offset the effect of the intervention on the money supply. If the central bank sells dollars and buys pesos, the money supply in pesos will increase while the money supply in dollars will decrease. This will cause the value of the peso to appreciate relative to the dollar, maintaining the current exchange rate.
\\\hline
         \textbf{GPT-3.5-turbo}& Based on the content provided, the information necessary to answer the question is not present. The content does not mention the central bank of Colombia or provide any information about the exchange rate of the Colombian currency. Therefore, we cannot determine which type of foreign market intervention the central bank of Colombia must conduct to keep the exchange rate at a level where the currency is not under- or overvalued in terms of PPP.
\\\hline
         \textbf{Ground Truth}& To eliminate the overvaluation in terms of PPP, the exchange rate for the Columbian peso needs to decline.  The central bank of Colombia should undertake an unsterilized foreign market intervention in which it will sell the domestic currency (COP) and buy foreign reserves, in order to increase reserves of its banking system, decrease the domestic interest rate, and shift the expected return on domestic currency denominated assets curve to the left.\\
    \hline
    \end{tabular}
    \caption{An Example of Unanswered Questions. We compare answers generated by GPT-3.5-turbo and Baichuan2-7B with the same embedder, retriever, and Reranker.}
    \label{tab:uaq}
\end{table*}

\end{document}